\pdfoutput=1
\documentclass[11pt]{article}

\usepackage[preprint]{acl}
\usepackage[ruled, linesnumbered, longend, nofillcomment]{algorithm2e}
\usepackage{listings}
\usepackage{multirow}
\usepackage{makecell}
\usepackage[para,online,flushleft]{threeparttable}
\usepackage{enumerate}
\usepackage{enumitem}
\usepackage{booktabs}
\usepackage{arydshln}
\usepackage{times}
\usepackage{latexsym}
\usepackage{amsmath}
\usepackage[T1]{fontenc}
\usepackage[utf8]{inputenc}

\usepackage{microtype}

\usepackage{inconsolata}

\usepackage{graphicx}
\newcommand{\model}{\textsc{AndroidGen}\xspace}

\newcommand{\planning}{ReflectPlan\xspace}
\newcommand{\autocheck}{AutoCheck\xspace}
\newcommand{\expsearch}{ExpSearch\xspace}
\newcommand{\gcevaluator}{StepCritic\xspace}

\newcommand{\gptfour}{GPT-4\xspace}

\newcommand{\gptfouro}{GPT-4o\xspace}
\newcommand{\llama}{Llama-3-70B\xspace}

\newcommand{\vpara}[1]{\vspace{0.04in}\noindent\textbf{#1}\xspace}
\newcommand{\hide}[1]{}

\title{\model: Building an Android Language Agent under Data Scarcity}

\author{
Hanyu Lai$^{1*\dagger}$, Junjie Gao$^{2*\dagger}$, Xiao Liu$^{1,2*}$, Yifan Xu$^{1\dagger}$, Shudan Zhang$^{1\dagger}$, \\{\bf Yuxiao Dong$^{1}$, Jie Tang$^{1\ddagger}$}\\
\\
\textsuperscript{1} Tsinghua University \quad
\textsuperscript{2} Zhipu AI
}

\begin{document}
\maketitle

\renewcommand{\thefootnote}{\fnsymbol{footnote}}
    \footnotetext[1]{LH, GJ and LX contributed equally.}
    \footnotetext[2]{Work done while these authors interned at Zhipu AI.}
    \footnotetext[3]{Corresponding author.}
\renewcommand{\thefootnote}{\arabic{footnote}}

\begin{abstract}
Large language models have opened up a world of possibilities for various NLP tasks, sparking optimism for the future. 
Despite their potential, LLMs have yet to be widely used as agents on real mobile devices.
The main challenge is the need for high-quality data sources.
Time constraints and labor intensity often hinder human annotation.
On the other hand, existing LLMs exhibit inadequate completion rates and need a robust data filtration strategy.
Given these challenges, we develop a framework called \model to enhance the capabilities of LLM-based agents under data scarcity. 
In addition, we leverage \model to collect trajectories given human tasks and train open-source LLMs on these trajectories to develop an open-source mobile agent without manually labeled trajectories.
We extensively evaluate \model with AndroidWorld, AitW, and various popular applications, demonstrating its improvements and revealing potential areas for future improvement.
Code, model, and data are available at \url{https://github.com/THUDM/AndroidGen}.
\end{abstract}

\begin{figure}[h!]
     \centering
   \includegraphics[width=\linewidth]{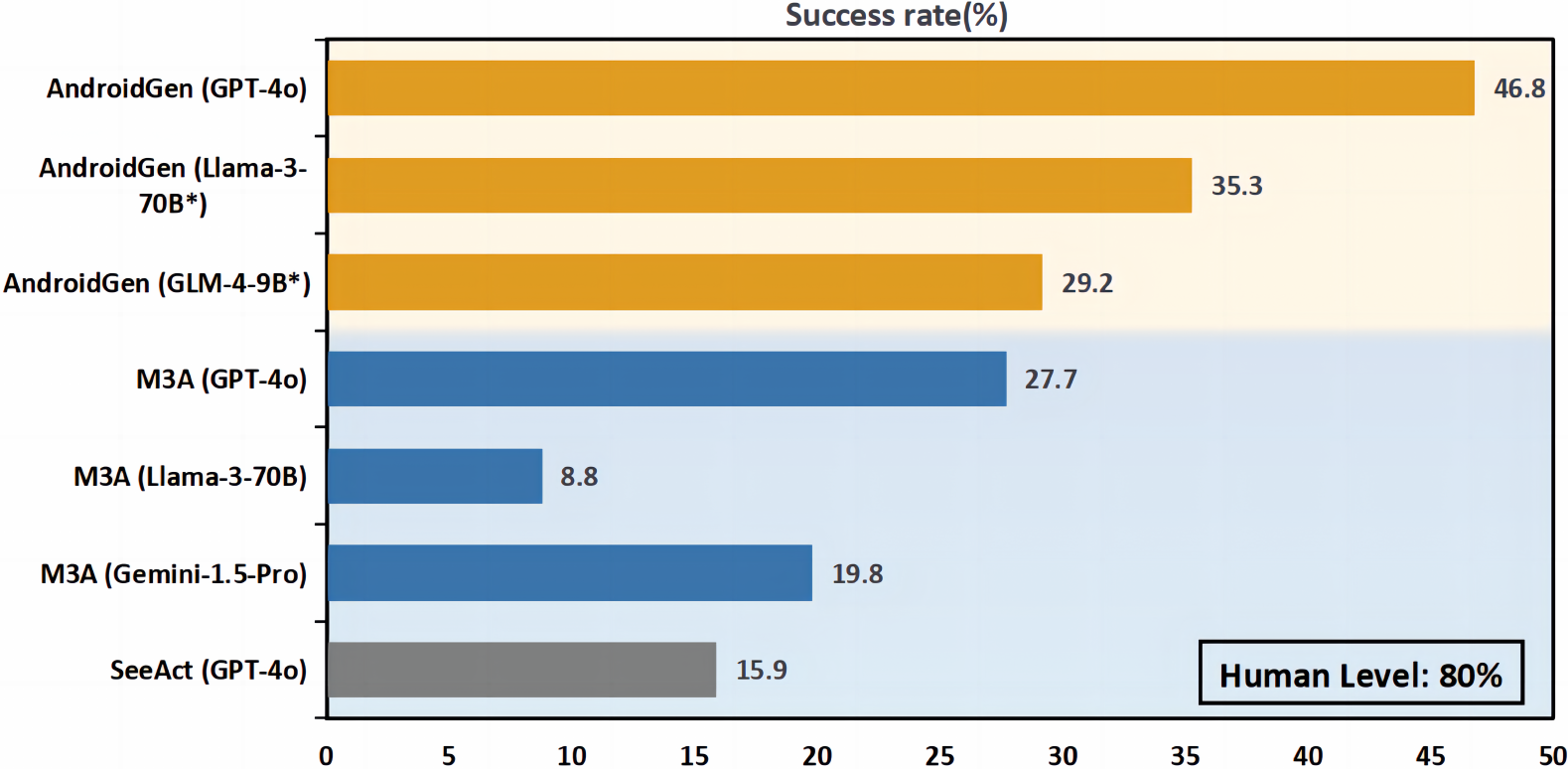}
    \vspace{-6mm}
   \caption{The success rates of popular mobile agents and humans on AndroidWorld.}
   \vspace{-8mm}
   \label{fig:success_rate}
\end{figure}

\begin{figure*}[t!]
     \centering
   \includegraphics[width=0.93\linewidth]{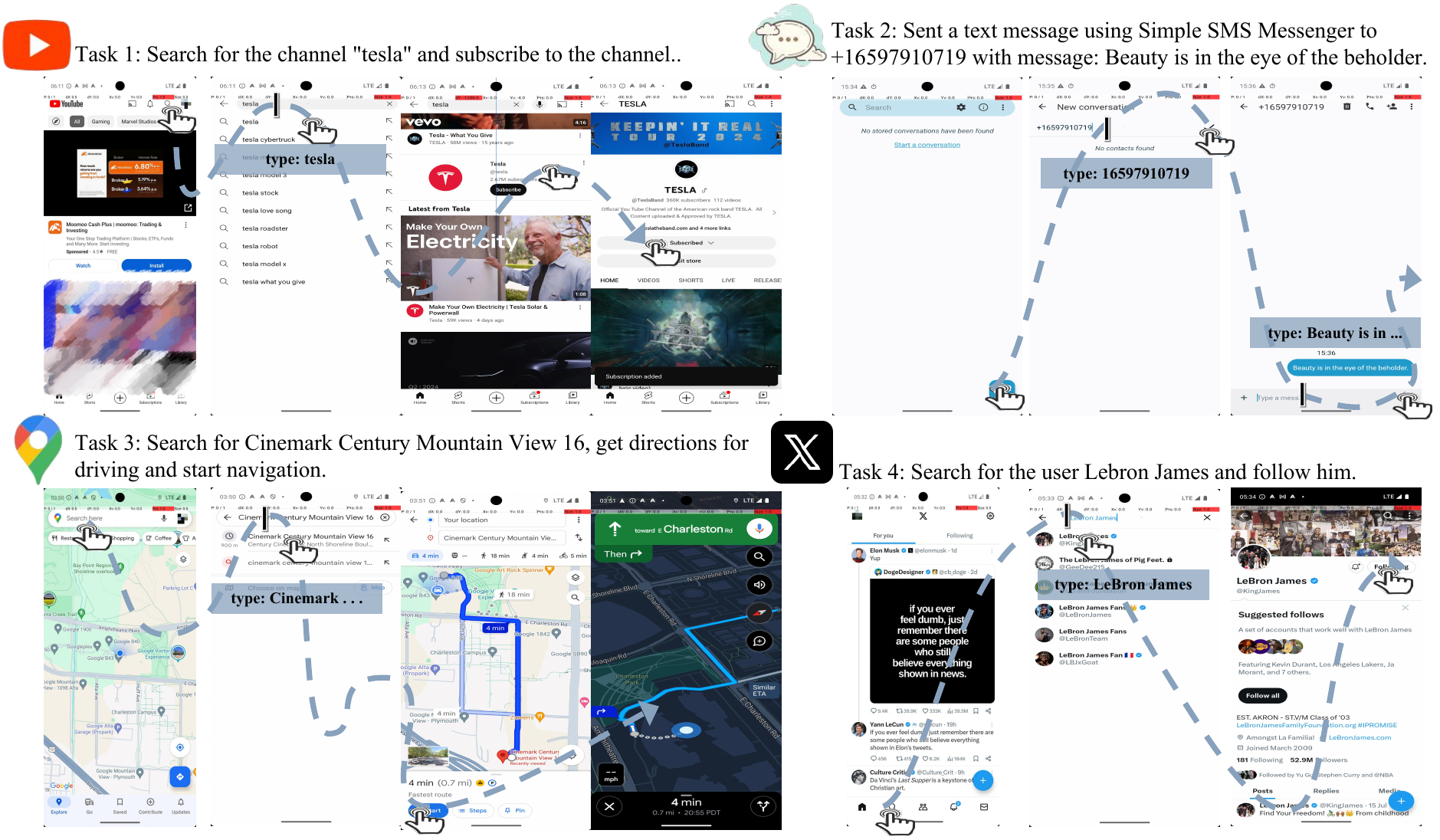}
    \vspace{-3mm}
   \caption{Examples of \model's execution on four user tasks.}
   \vspace{-5.5mm}
   \label{fig:demo}
\end{figure*}

\section{Introduction}
With the advancements in large language models (LLMs)~\cite{achiam2023gpt, touvron2023llama, touvron2023llama2, glm2024chatglm, zeng2022glm, zhang2022opt, scao2022bloom, team2023gemini}, the capabilities of AI have significantly expanded, reshaping our understanding of it.
These developments have heightened expectations for LLMs to act as intelligent agents and autonomously handle various tasks.
The emergence of mobile agents, primarily through pervasive smartphones, symbolizes a significant shift,  revolutionizing human-technology interactions and extending the boundaries of machine-driven productivity~\cite{xi2023rise, wang2023survey, liu2023agentbench}.

However, despite various attempts at mobile agents have achieved considerable success~\cite{baechler2024screenai, chen2024octopus, cheng2024seeclick, hong2023cogagent}, practical applications still face challenges. 
For instance, agents often fail when encountering complex tasks or unfamiliar scenarios.
Additionally, collecting extensive data on digital agents in real environments remains a significant unresolved challenge. 
Unlike traditional conversational datasets, data collection for digital environments presents the following difficulties:

\begin{itemize}
[leftmargin=*,itemsep=0pt,parsep=0.2em,topsep=0.2em,partopsep=0.0em]
\item \textbf{Scenario Diversity:} Different scenarios exhibit substantial variability, posing significant challenges to LLMs' generalization capabilities~\cite{lai2024autowebglm}. Consequently, the collection of tasks must encompass a wide range of diverse scenes and functionalities.
\item \textbf{Complex Task Data Collection:} For complex tasks involving multiple requirements, data collection typically entails numerous steps, necessitating robust planning abilities and precise execution~\cite{zhou2023webarena, rawles2024androidworld}. This process often leads to increased costs and lower completion rates.
\item \textbf{Data Filtration:} Effective data quality control demands meticulously examining the environments and operations to ensure full compliance with the task description. This process is both challenging and time-consuming, thereby further augmenting overall expenditures.
\end{itemize}

Currently, prevalent methods of manual and automated data collection~\cite{wang2022self, honovich2022unnatural, peng2023instruction, mukherjee2023orca} face significant challenges.
Manual annotation requires considerable time and financial resources, making collecting large volumes of high-quality trajectory data challenging.
On the other hand, utilizing advanced LLMs to accomplish tasks automatically is a potential method~\cite{xu2023wizardlm, luo2023wizardcoder, meng2022generating, lai2024autowebglm}. 
However, even state-of-the-art LLMs like \gptfour~\cite{achiam2023gpt} and Gemini~\cite{team2023gemini} still result in an unacceptably low success rate. Moreover, no effective automated solution selects high-quality, successful outcomes.

Inspired by these challenges, we build an agent framework, \model, designed to enhance the agent capabilities of LLMs in Android environments, particularly effective in scenarios where high-quality training data is scarce.
\model includes four modules: \expsearch, \planning, \autocheck, and \gcevaluator.
\begin{itemize}
[leftmargin=*,itemsep=0pt,parsep=0.2em,topsep=0.2em,partopsep=0.0em]
\item \expsearch enables LLMs to perform in-context learning through completed similar trajectories, thereby improving agent capabilities and facilitating generalization from simpler tasks to more complex ones through these samples.
\item \planning enables self-reflection on the current environment and updates the plan's status, thus enhancing the agent's long-term reasoning capabilities.
\item \autocheck proactively verifies the validity of each agent's operation, mitigating the risk of task failure due to operation errors.
\item \gcevaluator decomposes tasks into sub-goals and provides step-by-step trajectories evaluation, offering fine-grained labels for model optimization.
\end{itemize}

We leverage \model and LLMs to construct a robust Android agent under data scarcity. It can also serve as a pipeline to generate extensive browsing trajectories without human annotation. Furthermore, we introduce a data algorithm utilizing the fine-grained labels from \gcevaluator to filter and augment the data, thus creating a high-quality dataset for Android navigation agents. By fine-tuning open-source LLMs with this dataset, we develop a strong open-source Android agent without manually labeled trajectories.

To validate its effectiveness, we test \model on various Android benchmarks, including AndroidWorld~\cite{rawles2024androidworld}, AitW~\cite{rawles2023aitw}, and our popular Android app benchmark. 
The results demonstrate the design advantages of \model, particularly regarding reasoning capabilities, operational accuracy, and generalization ability. These findings underscore the potential of \model as a versatile and efficient tool in the mobile application.

In summary, our contributions are as follows:
\begin{itemize}[leftmargin=*,itemsep=0pt,parsep=0.2em,topsep=0.2em,partopsep=0.0em]
\item We develop \model, a novel Android agent framework, including \expsearch, \planning, and \autocheck to enhance the agent's reasoning capabilities and operation accuracy and empower it to generalize to complex tasks.
\item We introduce \gcevaluator for mobile devices providing fine-grained agent trajectories evaluation.
\item We employ \model to generate extensive trajectories and propose a data algorithm to construct a high-quality dataset. We then train open-source LLMs on this dataset to develop a robust open-source Android language agent.
\item We conduct evaluations of \model against several baselines across various Android benchmarks, demonstrating the improvements of \model over existing systems and identifying potential avenues for future research.
\end{itemize}
\section{Related Work}

\vpara{Large Language Models.}
Large language models (LLMs), such as GPT-4~\cite{achiam2023gpt}, Gemini~\cite{team2023gemini}, Claude-2~\cite{claude2}, the Llama series~\cite{touvron2023llama}, the ChatGLM series~\cite{zeng2022glm,du2022glm}, OPT~\cite{zhang2022opt}, and BLOOM~\cite{scao2022bloom}, have demonstrated remarkable capabilities in knowledge and language understanding, sparking a lot of research interest.

\vpara{In-Context Learning.}
LLMs have shown emergent abilities such as in-context learning~\cite{brown2020language,schick2020few} when the scale expands to a certain level. 
APE~\cite{zhou2022large} introduces an iterative search methodology to optimize prompts autonomously.
EPR~\cite{rubin-etal-2022-learning} retrieves relevant examples from a fixed dataset and incorporates these examples into the prompt to enhance response quality. These approaches leverage manual or static searching to perform in-context learning, restricting dynamic self-improvement within interactive environments.

\vpara{Mobile Agents.}
With LLMs continually surpassing expectations, numerous efforts have integrated them within digital environments~\cite{nakano2021webgpt, liu2023webglm, mei2024llm}.
Mobile phones represent typical environments for integration, given their pervasive role in daily life.
AppAgent~\cite{yang2023appagent} develops a multimodal framework allowing smartphones to learn and perform complex tasks through human-like interactions. 
Mobile-Agent~\cite{wang2024mobile} autonomously navigates and operates within app interfaces using visual perception, adapting across various mobile environments without XML. 
SeeAct~\cite{zheng2024gpt} leverages GPT-4V to act upon human instructions on websites, significantly surpassing text-only models when manually grounded. 
These agents rely on closed-source LLMs, often overlooking crucial aspects such as affordability, reproducibility, and transparency in complex interactive tasks.

\vpara{Machine Reasoning and Planning.}
To qualify as a robust agent, one must possess strong reasoning capabilities. 
The dual-system theory~\cite{daniel2017thinking} shed light on the cognitive processes in human thinking.
Chain-of-Thought~\cite{wei2022chain} enables LLMs to think, enhancing their reasoning abilities. 
ReAct~\cite{yao2022react} leverages LLMs to produce reasoning processes and actions, facilitating greater synergy. 
Several works~\cite{yin-etal-2024-agent,wang2023planandsolveprompting,xie2023translatingnatural} focus on task decomposition through subgoal generation with LLMs, improving the planning capabilities. 
However, these planning techniques do not reflect the plan's progress based on the environment and execution outcomes. Consequently, they may not be well-suited for complex multi-round scenarios.

\vpara{Benchmarks of Mobile Agents.}
The primary Android benchmarking approach relies on Android emulators, where agents attempt to complete various daily user tasks. 
AITW~\cite{xing2024understanding} provides a training dataset and offline evaluation metrics, utilizing automated partial match calculations to assess trajectory correctness. 
AndroidArena~\cite{xing2024understanding}, on the other hand, offers an emulator-based evaluation environment and employs the longest common subsequence to measure task completion rates. AndroidWorld~\cite{rawles2024androidworld} provides automated evaluation metrics for each task, resulting in a more precise task completion determination.

\vpara{Autonomous Evaluation.}
Pan et al.~\cite{pan2024autonomous} and DigiRL~\cite{bai2024digirl} propose using an LLM-based autonomous evaluator to assess task completion. However, their evaluation methodology is limited to binary outcomes and lacks adaptive fine-grained assessment of trajectories. They potentially overlook valuable trajectories, particularly in complex scenarios.

    \begin{figure*}[t]
        \centering
        \includegraphics[width=0.95\textwidth]{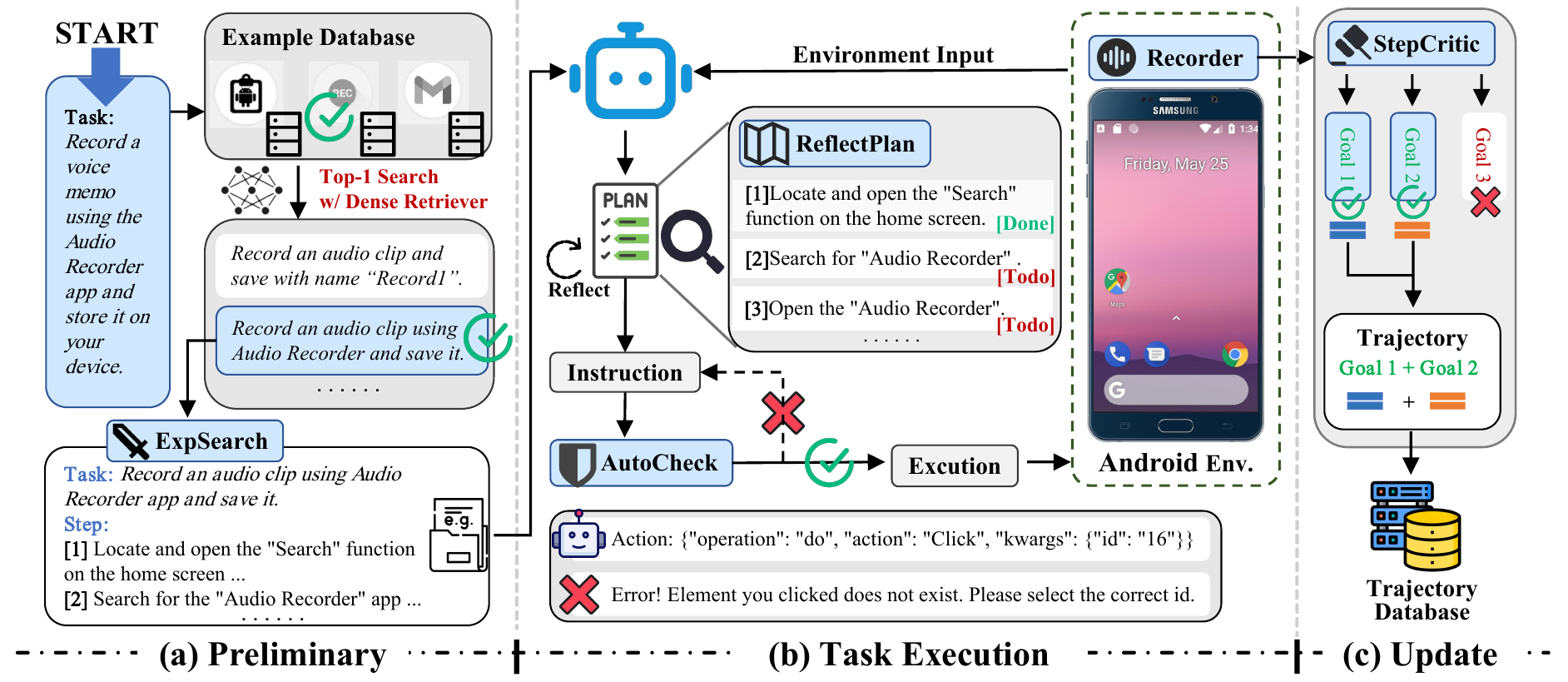}
        \vspace{-3mm}
        \caption{Overview of \model framework designed to complete tasks in Android. \textmd{Our process comprises three stages: preliminary, task execution, and update. Preliminary (a): \expsearch retrieve the top-1 similar tasks and trajectories from the database and feed them into the agent. Task Execution (b): \planning assesses the progress and updates the plan. Then, the agent generates operations based on the environment, plan, and retrieval example. \autocheck verifies these operations, executing them if successful or regenerating them if not. Update (c): \gcevaluator evaluates the trajectories in fine-grand and updates the database accordingly.}}
        \label{figs:main}
        \vspace{-5mm}
    \end{figure*}
    
    \section{\model Framework}\label{framework}
    This section introduces the \model framework for integrating LLMs into Android devices.
    It enables a language agent to complete user tasks through predefined input-output interfaces. 
    We also introduce a modular agent architecture to enhance the LLM's capabilities.
    
    \subsection{Environment}\label{environment}
    The environment, which serves as the substrate for agent interaction, is pivotal in determining the limit of task completion capabilities. We introduce our environment setup based on input and output.
    
    \subsubsection{Observation Space}\label{observation_space}
    Our observation space is designed to provide comprehensive and accurate inputs to our agents, ensuring that the received information mirrors what a human user would perceive. 
    We employ the Android XML as our environmental representation, simplifying and adding attribute details to facilitate element type determination. 
    This structured representation enhances the LLM's understanding of the environment, enabling it to grasp element attributes and states. The XML format can be found in Appendix~\ref{apdx:observation_space}.
    
    \subsubsection{Action Space}\label{action_space}
    The action space defines how the agent interacts with the environment. Establishing a complete action space is crucial for the agent to fulfill user requests. 
    We define our action space using Python function calls, leveraging the familiarity of LLMs with Python. Furthermore, we represent the action space using Python docstrings for brevity. 
    Based on our empirical insights, we delineate a concise yet extensive action space suitable for the Android environment, which can be found in Appendix~\ref{apdx:action_space}.
    
    \subsection{Agent Architecture}\label{agent_algorithm}
    After the environment construction, we design the agent architecture tailored to the digital environment. 
    We aim to increase the LLMs' capabilities under data-scarce conditions within the Android environment.
    The algorithmic workflow is illustrated in Figure~\ref{figs:main}. 
    Next, we will detail the implementation of the four modules: \expsearch, \planning, \autocheck, and \gcevaluator.
    
    \subsubsection{\expsearch}\label{expsearch}
    \expsearch is a novel approach leveraging LLM's in-context learning ability to optimize the agent iteratively by learning from its own trajectories. Our implementation consists of two parts:
    
    \vpara{Trajectory Collection}
    We gather trajectories where the agent self-samples to collect extensive data for the agent to learn from itself. 
    However, the challenge is ensuring these trajectories meet the task requirements.
    We employ \gcevaluator (see Section~\ref{judger}) to assess the trajectories based on their content and the given task.
    We preserve all the agent's trajectories and the completion status assessed by \gcevaluator, which will be utilized to construct our trajectory database (see Section~\ref{data_collection} for the details of the data collection process).
    
    \vpara{Trajectory Retrieval}
    Our next challenge is choosing the most similar and informative trajectory for the given task to let the agent learn from. 
    We first identify a collection within the same context, such as a specific application. 
    It is crucial as tasks, even when identical, may vary significantly in execution across different contexts.
    We then utilize Contriever~\cite{izacard2021towards} to encode instructions and compute similarity scores with embeddings from the database. The top-1 result is selected as our learning example.
    In addition, each time the agent completes a task, we use \gcevaluator to assess the trajectory and log it to the database, which enables our agent to self-improve iteratively, fostering easy-to-hard generalization in deployment.
    
    \subsubsection{\planning}\label{planning}
    In real-world complex digital environments, previous planning strategies are often overly optimistic about the execution results and prone to failure due to a lack of proper assessment of the current progress.
    Therefore, we develop \planning that enables self-assessment of the progress of tasks during execution. This approach empowers the agent to enhance planning and reflecting capabilities.
    \planning operates in two phases, as shown in Figure~\ref{figs:main} (b):
    
    \vpara{Plan Initialization.}
    Before the first execution step, the agent analyzes the task and the environment to generate a step-by-step plan to guide the following task execution.
    
    \vpara{Plan Reflection.}
    From the second step onwards, the agent will reflect on the current progress and update the plan based on the task's progress. 
    The agent can also revise and create new plans when encountering a failed state or entering a loop, enhancing planning robustness.
    
    \subsubsection{\autocheck}\label{autocheck}
    LLMs' operations are not flawless, even when the plan is correct. They are like humans who may make typos or incorrect clicks. 
    However, unlike humans, LLMs struggle to detect and correct simple errors, which can lead to task failure. 
    Therefore, we design the \autocheck module to mitigate the weakness and enhance agent robustness. 
    Upon generating the operation, \autocheck proactively verifies the response's validity. 
    When detecting potential issues or non-compliant actions, the subsequent execution is terminated, and feedback is provided to the agent in the next round.
    
    Our experiments show that self-checking operations cause inconsistent assessment standards, leading to false positives that can harm performance.
    We adopt a more straightforward and effective strategy: checking if each operation type achieves the expected outcome, such as the existence of element IDs, type compliance, and scroll completion. 
    Table~\ref{tab:autocheck} details the operation validation types.
    
    \vspace{-3mm}
\begin{table}[t]
    \small
    \caption{\autocheck type for \model}
    \vspace{-3mm}
    \renewcommand\tabcolsep{2pt}
    \label{tab:autocheck}
    \resizebox{\columnwidth}{!}{
    \begin{tabular}{ll}
        \toprule
        Function & Check Type \\
        \midrule
        \texttt{open\_app(app\_name)} & If \{ app\_name \} exists in the device\\
        \texttt{quote(content)} & If \{ content \} is empty\\
        \midrule
        \texttt{do(action,id,text,dir)} & \\
        \texttt{action type:} & \\
        \texttt{Click} & If element \{ id \} is on the screen\\
        \texttt{Long Press} & If element \{ id \} is on the screen\\
        \texttt{Input Text} & If element \{ id \} is on the screen\\
        & If element \{ id \} contains \{ text \}\\
        \texttt{Navigate Home} & If return to the home screen\\
        \texttt{Scroll} & If direction \{ dir \} is valid\\
        \texttt{Swipe} & If direction \{ dir \} is valid\\
        \bottomrule
    \end{tabular}
    }
    \vspace{-2mm}
\end{table}

    \subsubsection{\gcevaluator}\label{judger}
    To collect high-quality trajectories for \expsearch and training, we build \gcevaluator.
    \gcevaluator is built on \gptfouro, can decompose tasks into various sub-goals, and evaluate the trajectory step-by-step. This approach enables a granular assessment of the trajectories, maximizing the data's learning value.
    
\begin{figure}[h!]
        \centering
        \includegraphics[width=1.02\linewidth]{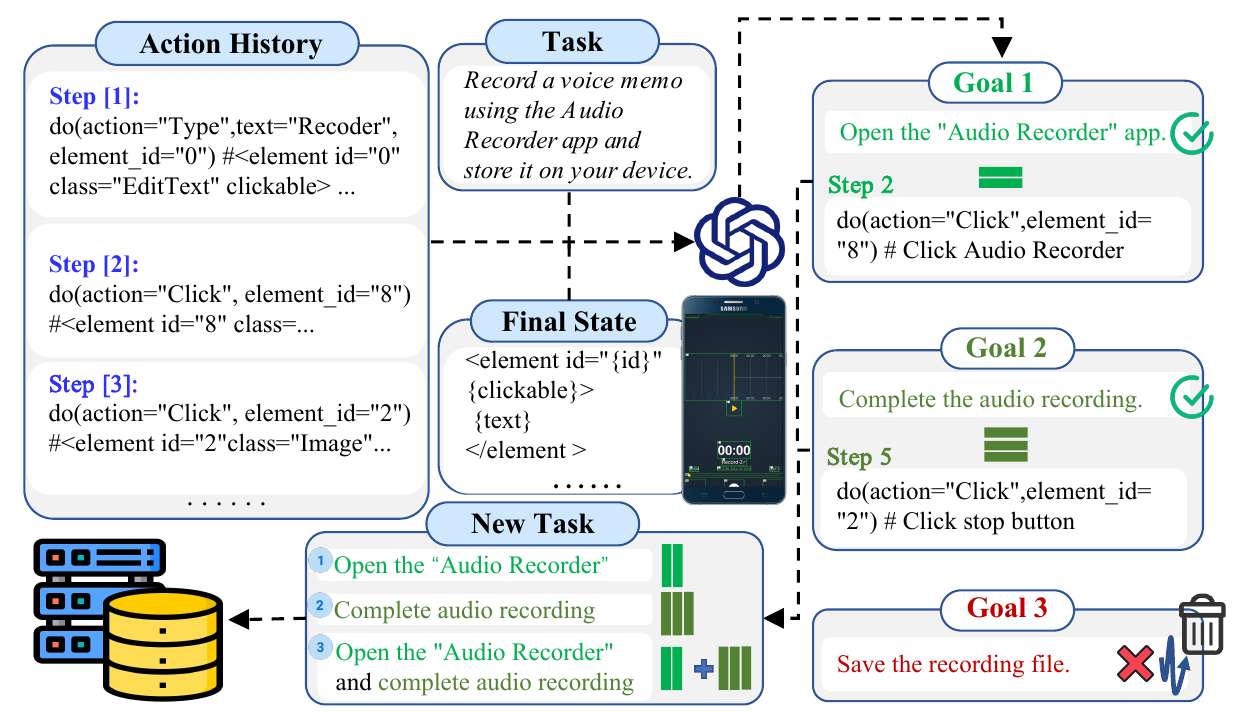}
        \vspace{-6mm}
        \caption{\model data construction workflow.}
        \label{figs:data_collection}
        \vspace{-3mm}
    \end{figure}

    Due to context length constraints, identifying critical information for evaluation is essential. 
    As shown in Figure~\ref{figs:data_collection}, we input the complete operation sequence, along with the final state of the device, to enhance the trajectory's information density under the constraint of limited context length. Then, we instruct \gcevaluator to assess whether each sub-goal is achieved and the corresponding steps.
    
    In addition to the environment input, we integrate the information from each module in the architecture to facilitate the agent's operations. For detailed prompt organization, refer to Appendix~\ref{apdx:agent_prompt}.
\section{Building An Open-Source Android Language Agent}
In Section~\ref{framework}, we develop \model framework, which integrates existing LLMs to serve directly as Android agents without prior training. 
Next, we will detail the data collection process and our designed data algorithm for \expsearch and training. 
Additionally, we will discuss our approach to model training on the synthesized Android trajectories, thereby constructing a robust open-source Android language agent without human annotation.

\subsection{Data Collection}\label{data_collection}
We demonstrate how to set up a pipeline for data construction with \model, as shown in Figure~\ref{figs:data_collection}.
This pipeline can efficiently generate a lot of high-quality Android browsing trajectories:

\vpara{Task Formulation.} We utilize \gptfouro to generate about 300 task instructions drawing on the instructions in AndroidWorld. We ensure no reward signals or golden labels are employed during training to prevent data leakage.

\vpara{Agent Sampling.} We then leverage \model with \gptfouro to sample a trajectory for each task.

\vpara{Trajectory Recording.} During the sampling process, we build a recorder that records environmental and operational information at each step, which is crucial for constructing a reproducible Android navigation trajectory. 

\vpara{Trajectory Evaluation.} Upon finishing each task, we utilize \gcevaluator to assess the recorded trajectories. \gcevaluator lists each sub-goal of the task along with the corresponding steps taken to achieve them (where -1 indicates incomplete). If each sub-goal is accomplished, the task is considered completed.

\vpara{Trajectory Augmentation.} For a task \( T \) and its corresponding sequence \( S \), \( T \) comprises multiple sub-goals \( g_1, g_2, \ldots, g_n \), each associated with completion steps \( p_1, p_2, \ldots, p_n \) (where \( p_i = -1 \) if the goal \( g_i \) is not completed). 
Let \( g_k \) be the first incomplete sub-goal (or \( g_n \) if all sub-goals are completed). 
We concatenate the sub-goals \( g_1 + \cdots + g_i \) for \( i \) ranging from 1 to \( k-1 \), and considering the subsequence \(\{p_1, \ldots, p_{k-1}\}\) as the label, to formulate new trajectories to augment our dataset. The algorithm pseudocode is in Appendix~\ref{apdx:algorithm}.

We integrate the original and augmented tasks from different sources to construct a dataset comprising more than 1000 trajectories.

\subsection{Training}
To develop a specialized open-source language agent, we fine-tune GLM-4-9B and \llama on automatically constructed datasets. This approach enables the creation of a robust Android agent without necessitating human annotation for trajectories.
We employ LoRA~\cite{hu2021lora} for LLM fine-tuning based on its advantages in requiring a lighter training load and strong model generalization, effectively mitigating overfitting.

We train the LLM with each step in the trajectories separately. 
To enhance deployment efficiency, we mix the planning and execution steps for fine-tuning, equipping the LLM with capabilities for both planning and execution. 
The prompts for training are consistent with those in Appendix~\ref{apdx:agent_prompt}, facilitating seamless integration of the trained model into \model framework.
For details of the training setting, refer to Appendix~\ref{apdx:training_details}.
\begin{table*}[t]
\caption{Per-task performance on AndroidWorld. \textmd{* indicates the model is fine-tuned.}}
\vspace{-3mm}
\label{tab:main_result}
\small
\renewcommand\tabcolsep{3pt}
\renewcommand\arraystretch{.92}
\resizebox{\textwidth}{!}{
\begin{tabular}{@{}l!{|}p{2cm}<{\centering}!{|}p{2cm}<{\centering}p{2cm}<{\centering}p{2cm}<{\centering}!{|}p{2cm}<{\centering}p{2cm}<{\centering}p{2cm}<{\centering}@{}}
\toprule
Agent  & SeeAct & M3A & M3A & M3A & \model & \model & \model \\
\midrule
Base Model & \gptfouro & Llama-3-70B & Gemini-1.5-Pro & \gptfouro & GLM-4-9B* & Llama-3-70B* & \gptfouro \\ 
\midrule
AudioRecorder & 50.0 & 50.0 & 50.0 & 50.0 & 100.0 & 100.0 & 100.0 \\
Calendar      & 0.0  & 0.0  & 5.9  & 11.8 & 0.0   & 6.3   & 12.5  \\
Camera        & 50.0 & 50.0 & 50.0 & 50.0 & 50.0  & 100.0 & 100.0 \\
Chrome        & 33.3 & 0.0  & 0.0  & 0.0  & 0.0   & 0.0   & 0.0   \\
Clock         & 33.3 & 33.3 & 66.7 & 66.7 & 33.3  & 66.7  & 66.7  \\
Contacts      & 0.0  & 0.0  & 33.3 & 33.3 & 0.0   & 33.3  & 66.7  \\
Expense       & 22.2 & 0.0  & 11.1 & 22.2 & 33.3  & 33.3  & 37.5  \\
Files         & 50.0 & 0.0  & 0.0  & 50.0 & 50.0  & 50.0  & 50.0  \\
Joplin        & 25.0 & 0.0  & 25.0 & 50.0 & 75.0  & 50.0  & 100.0 \\
Markor        & 7.1  & 0.0  & 7.1  & 14.3 & 14.3  & 28.6  & 35.7  \\
OpenTracks    & 0.0  & 0.0  & 0.0  & 0.0  & 16.7  & 16.7  & 16.7  \\
OsmAnd        & 33.3 & 0.0  & 33.3 & 33.3 & 33.3  & 0.0   & 33.3  \\
Recipe        & 8.3  & 0.0  & 18.2 & 25.0 & 41.7  & 40.0  & 45.5  \\
Retro         & 0.0  & 0.0  & 0.0  & 25.0 & 0.0   & 25.0  & 50.0  \\
Settings      & 26.7 & 26.7 & 50.0 & 57.1 & 64.3  & 70.0  & 93.3  \\
SMS           & 16.7 & 33.3 & 33.3 & 33.3 & 57.1  & 66.7  & 57.1  \\
Tasks         & 16.7 & 16.7 & 16.7 & 33.3 & 0.0   & 0.0   & 16.7  \\
Vlc           & 50.0 & 0.0  & 0.0  & 0.0  & 0.0   & 0.0   & 50.0  \\
\midrule
Avg.          & 15.9 & 8.8  & 19.8 & 27.7 & 29.2  & \underline{35.3}  & \textbf{46.8}  \\
\bottomrule
\end{tabular}
}
\vspace{-3mm}
\end{table*}

\section{Experiments}\label{experiment}
We conduct extensive experiments across various scenarios to evaluate the performance of the \model in executing a range of tasks within the Android digital environment compared to common baselines.
To closely reflect the real user experience, we select benchmarks that utilize interactive environments for evaluation. 
These benchmarks include AndroidWorld, AitW, and the benchmark we construct for popular applications.

\subsection{AndroidWorld}\label{exp:androidworld}
We evaluate \model on AndroidWorld \cite{rawles2024androidworld}, comparing our results with M3A \cite{rawles2024androidworld} and SeeAct \cite{zheng2024gpt}.
To ensure a fair comparison, we standardize the action space and a11y tree (text) as the environment input. 
We adapt SeeAct\textsubscript{choice} for the Android environment by augmenting the action space and applying heuristic filtering to modify the environment input, following the approach utilized in AndroidWorld.
We use task success rate as the evaluation metric. The results are in Table~\ref{tab:main_result}.
In addition, we perform a statistical analysis based on AndroidWorld's task difficulties to evaluate the performance on different levels. The results are in Appendix~\ref{apdx:difficulty_level}.

\subsection{Android in the Wild (AitW)}\label{exp:aitw}
We also evaluate \model on AitW~\cite{xing2024understanding}. 
To fairly compare DigiRL~\cite{bai2024digirl}, we follow its experiment setup and randomly select 96 tasks from the test splits provided in DigiRL. 
For the environment, we utilize the environment representation and action space defined in Section~\ref{environment} to evaluate \model. 
We leverage human experts for task assessments and compute the task success rate.
We compare \model with baselines in DigiRL, including prompting methods like Set-of-Mark~\cite{yang2023set} and AppAgent~\cite{yang2023appagent}, and training methods like SFT and Offline-to-Online RL~\cite{bai2024digirl}.
The results are in Table~\ref{tab:aitw}.

\begin{table}[t]
\caption{Performance on AitW. \textmd{* indicates the model is fine-tuned. Baselines are taken from DigiRL.}}
\label{tab:aitw}
\vspace{-3mm}
\centering
\renewcommand\tabcolsep{6pt}
\renewcommand\arraystretch{.95}
\resizebox{\columnwidth}{!}{
\begin{tabular}{lllcc}
\toprule
Method & & General & Web Shopping\\
\midrule
\multirow{2}{*}{Set-of-Mark} & Gemini-1.5-Pro & 13.5 & 8.3 \\
& \gptfouro & 16.7 & 11.5 \\
\hdashline
\multirow{2}{*}{AppAgent} & Gemini-1.5-Pro & 17.7 & 8.3 \\
& \gptfouro & 16.7 & 8.3 \\
\hdashline
\multirow{2}{*}{SFT} & CogAgent* & 25.0 & 38.5 \\
& AutoUI* & 14.6 & 17.7 \\
\hdashline
\multirow{2}{*}{Off-to-On RL} & Filtered BC* & 61.5 & 57.8 \\
& DigiRL* & 71.9 & 67.2 \\
\midrule
\multirow{3}{*}{\model} & GLM-4-9B* & 65.6 & 59.4 \\
& Llama-3-70B* & \underline{74.0} & \underline{79.2} \\
& \gptfouro & \textbf{85.4} & \textbf{81.3} \\
\bottomrule
\end{tabular}
}
\vspace{-3mm}
\end{table}

\subsection{Popular Applications}\label{exp:popular}
In addition to the reproducible benchmark, we select eight globally popular mobile applications, including Google Maps, X, YouTube, Spotify, Chrome, etc., for evaluation. 
We preinstall the applications on the simulator. For applications requiring login, we use a unified, pre-registered account to perform the login in advance. 
The environment settings for \model, SeeAct, and M3A follow the settings described in Section~\ref{exp:androidworld}.
For AppAgent, we employ screenshots with identifiers as input and its specific action space as output.
We construct five tasks for each application and employ human experts to judge the results and calculate the success rate.
The results are in Table~\ref{tab:appagent}.

\begin{table}[t]
\caption{Performance on popular applications. \textmd{* indicates the model is fine-tuned.}}
\label{tab:appagent}
\vspace{-3mm}
\centering
\renewcommand\tabcolsep{12pt}
\renewcommand\arraystretch{.95}
\resizebox{\columnwidth}{!}{
\begin{tabular}{llcc}
\toprule
Agent & Base Model & SR & Avg. Steps\\
\midrule
SeeAct & \gptfouro & 22.5 & 7.9 \\
M3A & \gptfouro & 40.0 & 7.3 \\
AppAgent & \gptfouro & \underline{57.5} & 6.7 \\
\midrule
\multirow{3}{*}{\model} & GLM-4-9B* & 35.0 & 7.2 \\
& Llama-3-70B* & 52.5 & 7.4 \\
& \gptfouro & \textbf{65.0} & 7.6 \\
\bottomrule
\end{tabular}
}
\vspace{-3mm}
\end{table}

\subsection{Evaluator Accuracy Comparison}\label{exp:evaluator}
To assess the performance of \gcevaluator, we employ trajectories generated by \model as the test set. We evaluate these trajectories using \gcevaluator and the baselines~\cite{pan2024autonomous}. 
We compare their prediction with environmental oracle prediction and compute the accuracy for entire trajectories to contrast the efficacy of \gcevaluator with that of the baselines. 
Moreover, we also manually evaluate \gcevaluator's accuracy in predicting whether each specific goal is achieved and the precision of the corresponding step prediction. 
The results are in Table~\ref{tab:judger}.

\begin{table}[t]
\centering
\caption{Evaluator accuracy on AndroidWorld.}
\vspace{-3mm}
\label{tab:judger}
\renewcommand\tabcolsep{8pt}
\renewcommand\arraystretch{.95}
\resizebox{\columnwidth}{!}{
\begin{tabular}{lccc}
\toprule
Model & \multicolumn{2}{c}{Sub-goal Acc.} & Overall Acc. \\
\cmidrule(lr){2-3}
& Completion & Step & \\
\midrule
Captioner + Mixtral & - & - & 82.4 \\
Captioner + GPT-4 & - & - & 84.6 \\
\midrule
\gcevaluator & 92.8 & 82.3 & \textbf{87.9} \\
\bottomrule
\end{tabular}
}
\vspace{-3mm}
\end{table}

\subsection{Ablation Study}
To assess the impact of different algorithms and training data on agent performance, we conduct a comprehensive ablation study in Table~\ref{tab:ablation_study}.
We conduct experiments on AndroidWorld as our primary performance indicator. To show the improvements more straightforwardly, we present the accuracy across different difficulty levels.

\begin{table}[t]
\caption{Ablation study.}
\label{tab:ablation_study}
\centering
\vspace{-3mm}
\renewcommand\tabcolsep{11pt}
\renewcommand\arraystretch{.95}
\resizebox{\columnwidth}{!}{
\begin{tabular}{@{}lcccc@{}}
\toprule
Method & Easy & Medium & Hard & Avg. \\ 
\midrule
\multicolumn{5}{c}{Strategy Ablation (w/ \gptfouro)}\\  
\midrule
Base Agent & 35.0 & 5.9 & 0.0 & 20.7 \\
+) \planning & 51.7 & 14.7 & 0.0 & 32.4 \\
+) \autocheck & 53.3 & 17.6 & 0.0 & 34.2 \\
+) \expsearch & \textbf{65.0} & \textbf{32.4} & \textbf{11.8} & \textbf{46.8} \\
\midrule
\multicolumn{5}{c}{Training Data Ablation (w/ Llama-3-70B)}\\ 
\midrule
Untrained & 18.3 & 2.9 & 0.0 & 10.8 \\
No Selection & 28.3 & 2.9 & 0.0 & 16.2 \\
Oracle-Selection & 48.3 & 2.9 & 0.0 & 27.0 \\
\gcevaluator & 43.3 & 5.9 & 0.0 & 25.2 \\
+) \expsearch & \textbf{55.8} & \textbf{12.5} & \textbf{5.9} & \textbf{35.3} \\
\bottomrule
\end{tabular}
}
\vspace{-5mm}
\end{table}

\vpara{Algorithm Ablation.}
For algorithm ablation, we validate the effectiveness of various algorithms on \gptfouro. 
The result indicates a 56.5\% overall SR improvement with \planning and 149.2\% for medium-difficulty tasks. 
Additionally, incorporating \autocheck reduces operational errors, increasing overall SR by 5.6\%. 
Lastly, \expsearch enables the agent to learn from simple tasks and generalize to more challenging ones, resulting in substantial improvements for both medium and hard tasks and an overall SR increase of 36.8\%.

\vpara{Training Data Ablation.}
For the ablation of data selection, we conduct experiments using \llama. 
We compare several scenarios: untrained, no selection (i.e., all data), and data selection. 
For selection methods, we choose \gcevaluator, environmental feedback (oracle), and \gcevaluator with \expsearch. 
The result shows that data selection with \gcevaluator significantly improves performance compared to the untrained and no-selection scenarios, approaching the efficacy of oracle selection. 
Integrating \expsearch enables our agent to generalize from examples, achieving a substantial improvement.

\subsection{Case Study and Error Analysis}
We conduct case studies on Android devices to explore potential optimizations. 
\model yields satisfactory results in most scenarios.
However, our agent also has limitations. We identify occasional errors during task execution, broadly categorized into four types: vision, mathematical counting, multiple app interactions, and memorization. Table~\ref{tab:error_distribution} summarizes the proportions of these error types during evaluation. These errors highlight areas for future improvement of our Android agent.
Detailed descriptions of specific good and bad cases are in Appendix~\ref{apdx:demo}.

\begin{table}[t]
\centering
\caption{Error distribution of \model.}
\vspace{-3mm}
\label{tab:error_distribution}
\renewcommand\tabcolsep{12pt}
\renewcommand\arraystretch{.95}
\resizebox{0.6\columnwidth}{!}{
\begin{tabular}{lc}
\toprule
Error Type & Proportion \\
\midrule
Vision & 15\% \\
Math Counting & 23\% \\
Multiple App & 26\% \\
Memorization & 20\% \\
Others & 16\% \\
\bottomrule
\end{tabular}
}
\vspace{-1mm}
\end{table}

\subsection{Efficiency \& Cost Analysis}
Last, we calculate the efficiency and cost of \model and M3A~\cite{rawles2024androidworld} in data construction and compare with human annotation (compliant with local regulations). 
Using 1,000 samples as an example, without quality control, \model's efficiency and cost are slightly better than those of M3A. Besides, our cost is 12.5\% of human annotation, and its efficiency is 275\%. 
For quality-controlled data generation (not compared with M3A due to the lack of quality control), \model achieves efficiency 5.85 times greater with only 5\% of the human annotation cost after selection and augmentation by \gcevaluator.

\begin{table}[t]
\centering 
\caption{Efficiency and cost statistics for M3A, human annotation, and \model (for 1,000 trajectories).}
\vspace{-3mm}
\label{tab:cost}
\renewcommand\tabcolsep{5pt} 
\renewcommand\arraystretch{.95} 
\resizebox{\columnwidth}{!}{
\begin{tabular}{lccc} 
\toprule 
Metrics & M3A & Human & \model \\ 
\midrule 
Average Time per Step (s) & 4.8 & 18 & 5.6 \\ 
Average Time per Task (s) & 49.5 & 120 & 43.7 \\ 
Cost per Task (\$) & 0.12 & 0.8 & 0.1 \\ 
\midrule 
\multicolumn{4}{l}{\textit{Without Quality Control:}} \\ 
Total Cost (\$) & 120 & 800 & \underline{100} \\ 
Total Time (hr) & 13.75 & 33.3 & \underline{12.1} \\ 
\midrule 
Success Rate (\%) & 32 & 80 & 58.2 \\ 
Evaluation Cost per Task (\$) & - & 0.20 & 0.005 \\ 
Evaluation Time per Task (s) & - & 30 & 3.5 \\ 
Data Augmentation (Factor) & - & - & 2.52 \\ 
\midrule 
\multicolumn{4}{l}{\textit{With Quality Control:}} \\ 
Total Cost (\$) & - & 1,250 & \textbf{71.6} \\ 
Total Time (hr) & - & 52.1 & \textbf{8.9} \\ 
\bottomrule 
\end{tabular}
}
\vspace{-3mm}
\end{table}

\section{Conclusion}
In this work, we present \model, a language agent framework for Android, including \expsearch, \planning, \autocheck, and \gcevaluator, significantly enhancing the agent's ability to perform complex tasks under data scarcity. 
We employ \model as a pipeline to efficiently construct training datasets and train a robust open-source Android language agent without human-annotated trajectories.
Our findings show significant progress in utilizing LLMs for the Android agent.
\clearpage

\bibliography{custom}

\appendix

\section{AndroidWorld with Difficulty Levels}\label{apdx:difficulty_level}

Table~\ref{tab:difficult} presents the performance of various agents across different difficulty levels in AndroidWorld.

\begin{table}[h]
\caption{Performance on various difficulty levels in Androidworld. \textmd{* indicates model is fine-tuned.}}
\label{tab:difficult}
\vspace{-3mm}
\centering
\renewcommand\tabcolsep{6.5pt}
\renewcommand\arraystretch{.95}
\resizebox{\columnwidth}{!}{
\begin{tabular}{llcccc}
\toprule
Agent & Base Model & Easy & Medium & Hard & Avg. \\
\midrule
SeeAct & \gptfouro & 28.3 & 2.9 & 0.0 & 15.9 \\
M3A & Llama-3-70B & 15.0 & 2.9 & 0.0 & 8.8 \\
M3A & Gemini-1.5 Pro & 33.3 & 5.9 & 0.0 & 19.8 \\
M3A & \gptfouro & 45.9 & 8.8 & 0.0 & 27.7 \\
\midrule
\model & GLM-4-9B* & 47.5 & 8.8 & 5.6 & 29.2 \\
\model & Llama-3-70B* & 55.8 & 12.5 & 5.9 & \underline{35.3} \\
\model & \gptfouro & 65.0 & 32.4 & 11.8 & \textbf{46.8} \\
\bottomrule
\end{tabular}
}
\vspace{-2mm}
\end{table}
\section{Training Details}\label{apdx:training_details}
During our LoRA fine-tuning of the GLM-4-9B and Llama-3-70B model, we employ a single-node eight-GPU A100-80B machine. The fine-tuned parameters account for about 0.024\% of the total parameters. We set the maximum learning rate to 1e-4 and the sequence length to 8192 and use a total batch size of 32, conducting the training over 3 epochs.
\section{Data Augmentation Algorithm Pseudocode}\label{apdx:algorithm}
We explain our data augmentation process as outlined in Algorithm~\ref{tab:algorithm}.

\begin{algorithm}
    \DontPrintSemicolon
    \SetAlgoLined
    \SetNoFillComment
    \caption{Data Augmentation Algorithm}
    \label{tab:algorithm}
    \KwData{Sequence of operations \( S \) for task \( T \) with sub-goals \( g_1, g_2, \ldots, g_n \) and completion steps \( p_1, p_2, \ldots, p_n \)}
    \KwResult{Augmented dataset with new trajectories}

    \For{each task \( T \)}{
        \For{each completion step \( p_i \)}{
            \If{\( p_i = -1 \)}{
                $k \gets i$ \;
                \textbf{break} \;
            }
        }
        \If{all sub-goals are completed}{
            $k \gets n$ \;
        }
        new\_trajectory $\gets \{\}$ \;
        \For{$i \gets 1$ \textbf{to} $k-1$}{
            concatenate $g_i$ to \( T_{\text{new}} \) \;
            add ( \( T_{\text{new}} \), \{ $p_1$, \ldots, $p_i$ \} ) to new\_trajectory \;
        }
        add new\_trajectory to augmented\_dataset \;
    }
    \Return{augmented\_dataset}
\end{algorithm}

\section{Action Space}\label{apdx:action_space}
Our action space for \model is shown in Table~\ref{tab:action_space}.

\begin{table}[htbp]
    \small
    \caption{Action space for \model}
    \vspace{-3mm}
    \renewcommand\tabcolsep{2pt}
    \label{tab:action_space}
    \resizebox{\columnwidth}{!}{
    \begin{tabular}{ll}
        \toprule
        Function & Description \\
        \midrule
        \texttt{open\_app(app\_name)} & Open the app with \{ app\_name \}\\
        \texttt{quote(content)} & Record \{ content \} for memory\\
        \texttt{exit(message)} & End the task with \{ message \}\\
        \midrule
        \texttt{do(action,id,text,dir)} & Do a single operation\\
        \texttt{action type:} & \\
        \texttt{Click} & Click on \{ id \} element\\
        \texttt{Long Press} & Long press on \{ id \} element\\
        \texttt{Input Text} & Input \{ text \} to \{ id \} element\\
        \texttt{Press Enter} & Press enter key\\
        \texttt{Navigate Home} & Return to the home screen\\
        \texttt{Navigate Back} & Return to the previous page\\
        \texttt{Scroll} & Scroll in \{ dir \}\\
        \texttt{Swipe} & Swipe \{ id \} element in \{ dir \}\\
        \texttt{Wait} & Wait for a while\\
        
        \bottomrule
    \end{tabular}
    }
    \vspace{-2mm}
\end{table}
\section{Observation Space}\label{apdx:observation_space}
We present the element format of the environment in our observation space as follows:
\lstset{
    backgroundcolor=\color[RGB]{245,245,244},
    breaklines=true,
    basicstyle=\ttfamily\small
}\begin{lstlisting}
<element id="{id}" class="{class}" resource-name="{resource}" {clickable/checkable/status="on"|"off"/editable}> {text} </element>
\end{lstlisting}
\section{Agent Prompt}\label{apdx:agent_prompt}
To enhance the model's understanding of the tasks and the environment, we design a set of model prompts, which are categorized into the following components:

\vpara{Setup}: In the setup, we articulate the objectives of the task and the rules for interacting with the environment. We utilize in-context learning to enable the LLM to act as an expert assistant within Android, stimulating the model's agent ability in the environment.

\vpara{Operation Definition}: We define the interfaces of interaction with the environment. 
We employ Python docstrings to define operation types, which are readily recognizable and understandable by most LLMs. The code-based approach promotes the model's reasoning capabilities and operational accuracy.

\vpara{Example}: The complete trajectory (omitting specific environment inputs) retrieved by \expsearch (in Section~\ref{expsearch}) will be used as a reference example to facilitate one-shot learning for the model. 
In addition, we include analysis and confirmation in the example to spur the model's reasoning and self-checking abilities.

\vpara{Note}: Based on our deployment practices, we include reminders to help agents avoid typical common errors in the note section.

\vpara{History}: We present the operational history in a conversational format involving the environmental input as the user and the operation output as the assistant, which is well-received by most LLMs. We have obscured historical environmental input information to reduce the context length significantly.

\vpara{Current Status}: Besides the environment states, we also incorporate the current plan and feedback from the last round (if available) as the current context for our agent.

We assemble the components above into a system prompt and model dialogue prompts, which are then inputted to the agent.
Below is our detailed prompt organization for \model:
\lstset{
    backgroundcolor=\color[RGB]{245,245,244},
    breaklines=true,
    basicstyle=\ttfamily\small
}\begin{lstlisting}
# Setup
You are a professional agent assistant that can fulfill user's high-level instructions on android devices. Given the current state of the device, you should first read carefully the plan that user provide, then generate operations to complete the todo goal in python-style pseudo code using the predefined functions.

# More details about the code
Your code should be readable, simple, and only **ONE-LINE-OF-CODE** at a time, avoid using loop statement and only use if-else control if necessary.

# Predefined functions are as follow:
```
def open_app(app_name):
	"""Open the app on the android device.

	Args:
		:param app_name: the name of the app to open.

	Returns:
		None. The app will be opened.
	"""

def do(action, element_id, text, direction):
	"""A single operation on the android device.

	Args:
		:param action: one of the actions from ["Click", "Long Press", "Input Text", "Press Enter", "Navigate Home", "Navigate Back", "Scroll", "Swipe", "Wait"]. "Swipe" is the inverse of "Scroll".
        :param element_id: optional. Only for ["Click", "Long Press", "Input Text", "Swipe"].
		:param text: optional. Only for ["Input Text"], indicating the text to input.
        :param direction: optional. Only for ["Scroll", "Swipe"], indicating the direction to scroll, choose from ["up", "down", "left", "right"].

	Returns:
		None. The device will be updated after executing the action.
	""	

def quote(content):
    """Quoting information from the current page for memory. Only you can see the quoted content.

    Args:
        :param content: text summarized or copied from the page for later operation.

    Returns:
        None.
    """

def exit(message):
	"""Ending the operation process if the assistant think it has fulfilled the goal.

	Args:
		:param message: optional. If user's instruction is a question, return assistant's answer in the message based on the operation result.

	Returns:
		None.
	"""
```

# A reference example:
{example_from_expsearch}

# REMEMBER: 
- Only **ONE-LINE-OF-CODE** at a time.
- You should follow the plan that user provide and do the operation step by step.
- Confirm and Analysis at the beginning of each round.
- You must ensure that the id of the element you act for is in the current page, or you shouldn't do acitions on the nonexistent element.
- If your action includes an element id, you should add a comment in the code to explain the element id.
- "Input Text" action will delete the original text in the input box and input the new text. You should concatenate the text if you want to add text to the original text.\end{lstlisting}

Below is our history and input prompt for \model:
\lstset{
    backgroundcolor=\color[RGB]{245,245,244},
    breaklines=true,
    basicstyle=\ttfamily\small
}\begin{lstlisting}
<|user|>
User Instruction: {user_instruction}

<|assistant|>
** Task Start **

<|user|>
** Environment State (Omitted) **

<|assistant|>
{round0_operation}

<|user|>
** Environment State (Omitted) **

<|assistant|>
{round1_operation}

<|user|>
** Environment State (Omitted) **

<|assistant|>
{round2_operation}

...

<|user|>
# Current State: {current_environment_text}

# Plan: {current_plan}

# Last Round Error: {error_feedback}\end{lstlisting}
\section{Evaluator Prompt}\label{apdx:evaluator_prompt}
Below is the system prompt for our \gcevaluator:
\lstset{
    backgroundcolor=\color[RGB]{245,245,244},
    breaklines=true,
    basicstyle=\ttfamily\small
}\begin{lstlisting}
You are an expert in evaluating the performance of an android agent. The agent is designed to help a human user navigate on their device to complete a task. Given the user's intent, the agent's action history, the final state of the device, and the agent's response to the user, your goal is to list the conditions the task requires and their respective completion step (or -1 for not completed). There are two types of tasks:
1. Information seeking: The user wants to obtain certain information from the screen, such as information about a product, reviews, map info, comparison of map routes, etc. The bot's response must contain the information the user wants or explicitly state that the information is not available. Otherwise, e.g., if the bot encounters an exception and responds with the error content, the task is considered a failure. Besides, be careful about the sufficiency of the agent's actions. For example, when asked to list the top-searched items in a shop, the agent should order the items by the number of searches and then return the top items. If the ordering action is missing, the task is likely to fail.
2. Application Operation: The user wants to do operations in a specific application, such as purchasing, modifying content or configuration, starting a project, etc. Carefully examine the bot's action history and the final state of the page to determine whether the bot completes the task. No need to consider the bot's response.

*IMPORTANT* Your output should STRICTLY follow the format below and DONOT output other content:
```
"condition1": "completion step1 (or -1 for not completed)"
"condition2": "completion step2 (or -1 for not completed)"
...
```\end{lstlisting}

For the input prompt, we provide the user's task, action history, and the final state of the screen. The LLM then evaluates the trajectory based on the information provided.
\section{Demonstration}\label{apdx:demo}
This section demonstrates examples from diverse applications, encompassing the first eight good cases and the last three bad cases. These instances embody distinct operational logic and require different functionalities to solve the challenges and requirements encountered when engaging with various applications.

\subsection{Audio Recorder}
\begin{figure*}[h]
     \centering
   \includegraphics[width=0.95\linewidth ]{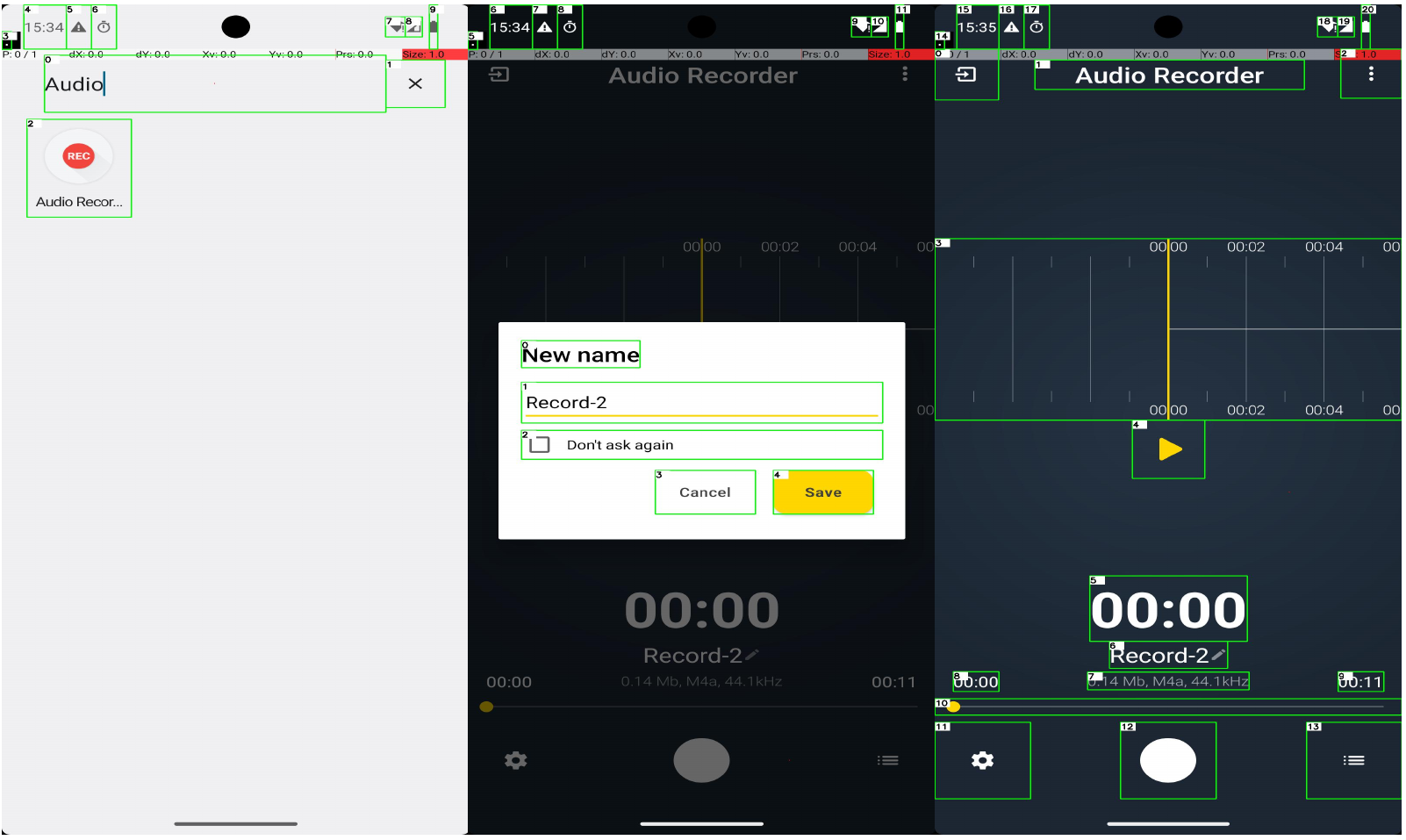}
    \vspace{-4mm}
   \caption{Audio Recorder}
   \vspace{4mm}
   \label{fig:demo1}
\end{figure*}

The targeted task to be executed is "Record an audio clip using the Audio Recorder app and save it." The actual execution steps can be summarized as follows:
\begin{itemize}
    \item Step1: Click SearchBar
    \item Step2: Type SearchBar "Audio Recorder"
    \item Step3: Click "Audio Recorder"
    \item Step4: Click "Start Recording"
    \item Step5: Click "End Recording"
    \item Step6: Save the file
\end{itemize}
As Figure~\ref{fig:demo1} shows, we end up with a recorded audio on the screenshot. We recorded a 12s audio recording, effectively completing the target task.

\subsection{Phone Contact}
\begin{figure*}[h]
     \centering
   \includegraphics[page=5,width=0.95\linewidth ]{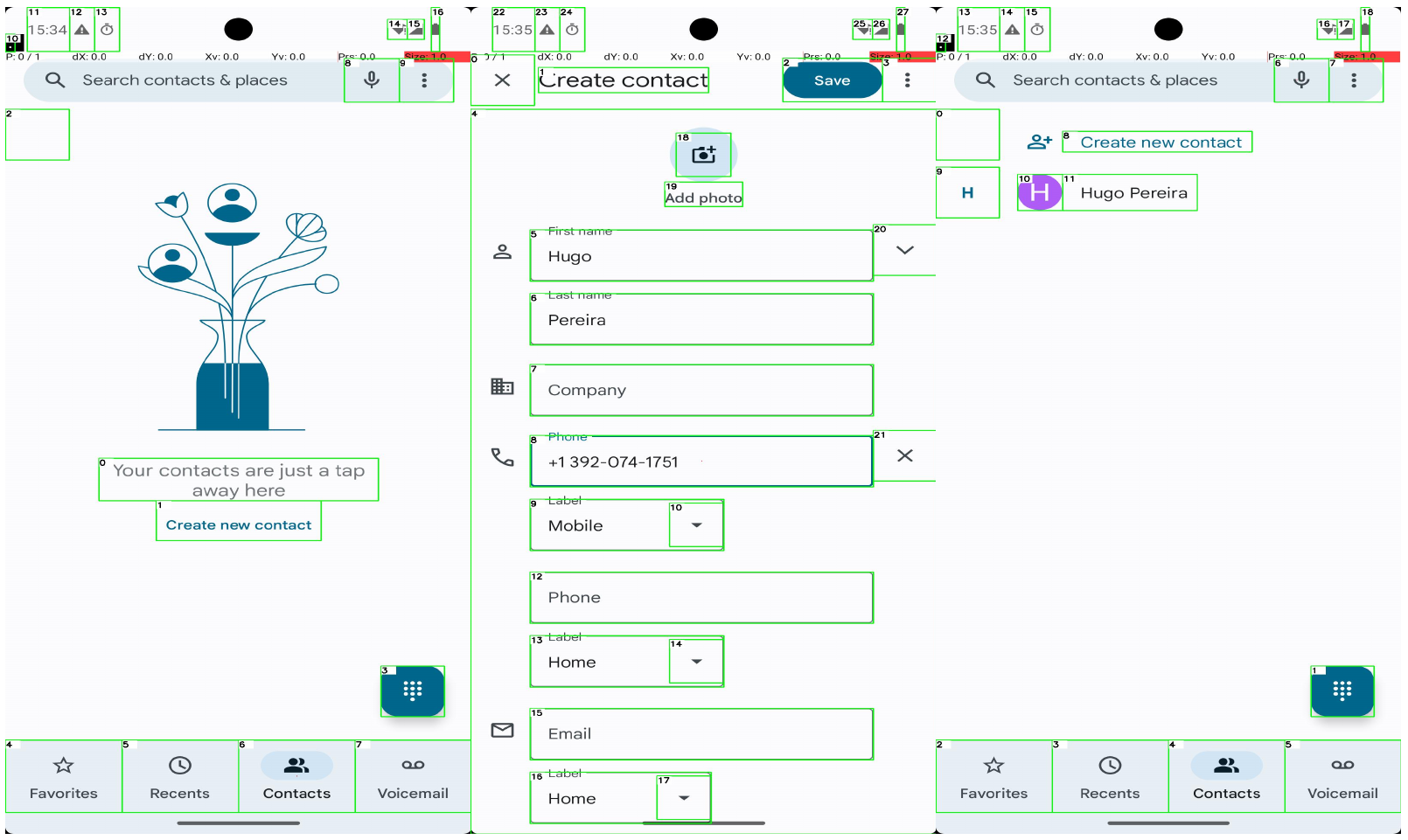}
    \vspace{-4mm}
   \caption{Phone Contact}
   \vspace{-4mm}
   \label{fig:demo2}
\end{figure*}

The targeted task is "Create a new contact for Hugo Pereira. Their number is +13920741751". The actual execution steps can be summarized as follows:
\begin{itemize}
    \item Step1: Click "Phone" App
    \item Step2: Click "Contacts"
    \item Step3: Click the "Add" button
    \item Step4: Type contact information 
    \item Step5: Save the contact
\end{itemize}
As Figure~\ref{fig:demo2} shows, we end up on the screenshot with a contact named Hugo Pereira. We add a contact with his name and phone number, effectively completing the target task.

\subsection{Delete File}
\begin{figure*}[h]
     \centering
   \includegraphics[page=1,width=0.95\linewidth ]{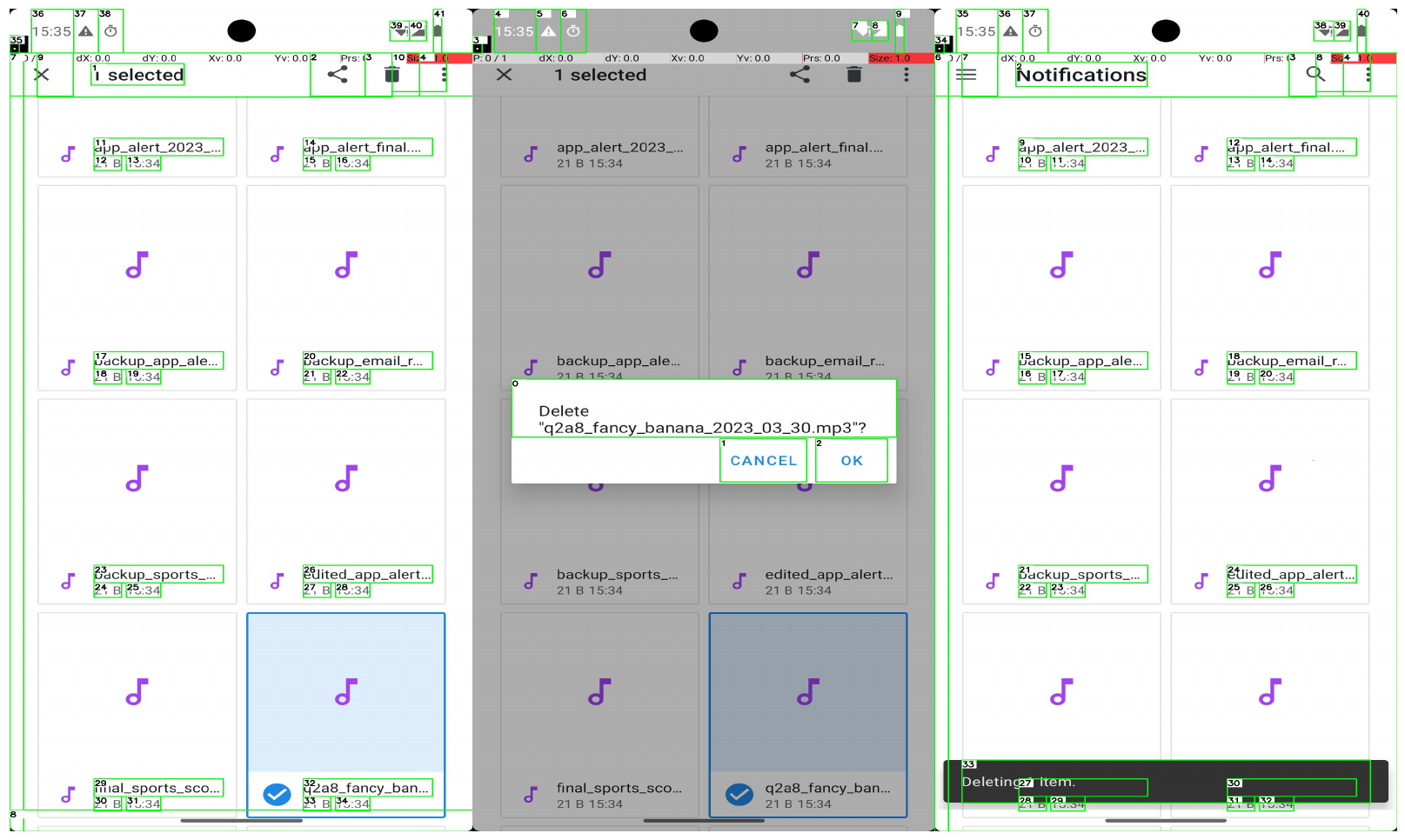}
    \vspace{-4mm}
   \caption{Delete File}
   \vspace{4mm}
   \label{fig:demo3}
\end{figure*}

The targeted task to be executed is "Delete the file banana.mp3 from the Android filesystem located in the Notifications folder within the sdkgphone storage area". The actual execution steps can be summarized as follows:
\begin{itemize}
    \item Step1: Open "Files" App
    \item Step2: Click "Show roots"
    \item Step3: Click "sdkgphone"
    \item Step4: Click "Notifications"
    \item Step5: Scroll Down
    \item Step6: Long Press "banana.mp3"
    \item Step7: Click "Delete"
    \item Step8: Click "OK"
\end{itemize}
As Figure~\ref{fig:demo3} shows, we end up on the screenshot in the Notifications folder. We delete the file banana.mp3, effectively completing the target task.

\subsection{Marking Map}
\begin{figure*}[h]
     \centering
   \includegraphics[page=4,width=0.95\linewidth ]{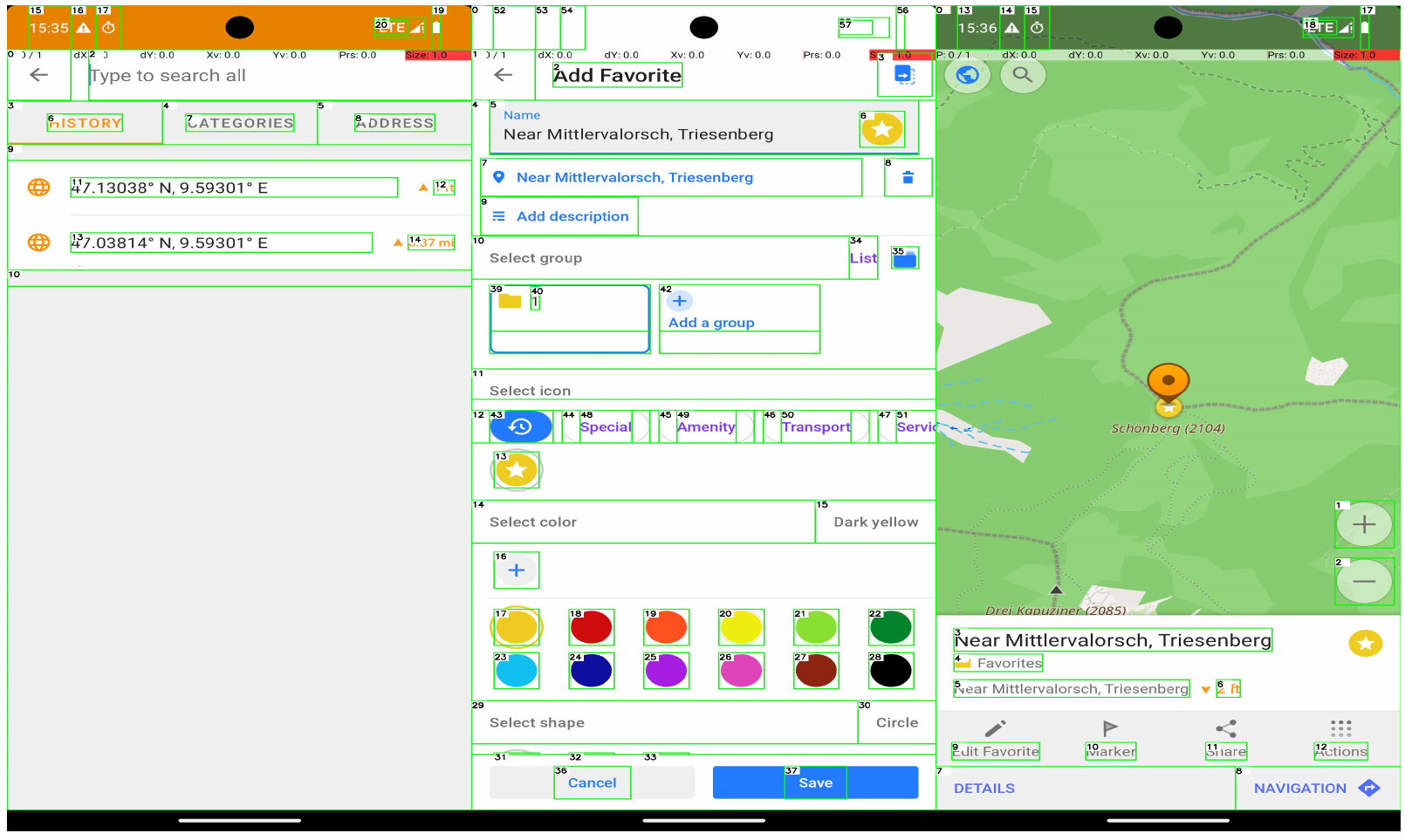}
    \vspace{-4mm}
   \caption{Marking Map}
   \vspace{-4mm}
   \label{fig:demo4}
\end{figure*}

The targeted task is "Add a favorite location marker for 47.1303814, 9.5930117 in the OsmAnd maps app". The actual execution steps can be summarized as follows:
\begin{itemize}
    \item Step1: Open app "OsmAnd" App
    \item Step2: Click the "Search" button
    \item Step3: Type "47.1303814, 9.5930117"
    \item Step4: Click "Show on map"
    \item Step5: Long Press
    \item Step6: Click the "Add" button
    \item Step7: Click the "Add Favorite" button
    \item Step8: Click the "Save" button
\end{itemize}
As Figure~\ref{fig:demo4} shows, we end up on the OsmAnd maps app with a location marker. We add a favorite location marker, effectively completing the target task.

\subsection{Create Music Playlist}
\begin{figure*}[h]
     \centering
   \includegraphics[page=2,width=0.95\linewidth ]{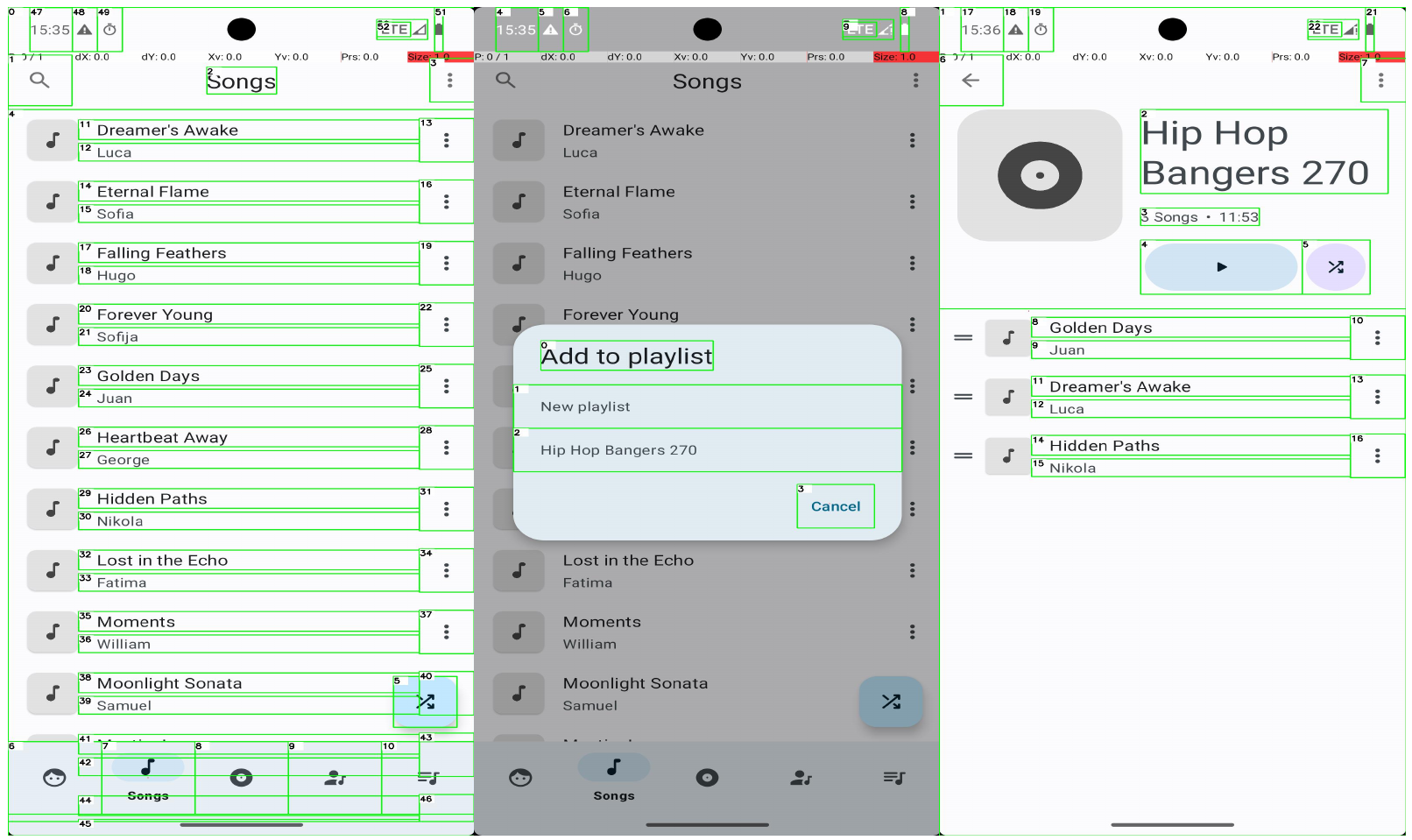}
    \vspace{-4mm}
   \caption{Create Music Playlist}
   \vspace{4mm}
   \label{fig:demo5}
\end{figure*}

The targeted task to be executed is "Create a playlist in Retro Music titled "Hip Hop Bangers 270" with the following songs, in order: Golden Days." The actual execution steps can be summarized as follows:
\begin{itemize}
    \item Step1: Open app "Retro Music" App
    \item Step2: Click "Playlist"
    \item Step3: Click "More Options"
    \item Step4: Click "New Playlist"
    \item Step5: Type "Hip Hop Bangers 270"
    \item Step6: Click "Create"
    \item Step7: Click "Songs"
    \item Step8: Long Press "Golden Days"
    \item Step9: Click "Add to playlist"
    \item Step10: Click "Hip Hop Bangers 270"
\end{itemize}
As Figure~\ref{fig:demo5} shows, we end up on the "Retro Music" playlist. We add "Golden Days" to the playlist, effectively completing the target task.

\subsection{Turn Bluetooth Off}
\begin{figure*}[h]
     \centering
   \includegraphics[page=3,width=0.95\linewidth ]{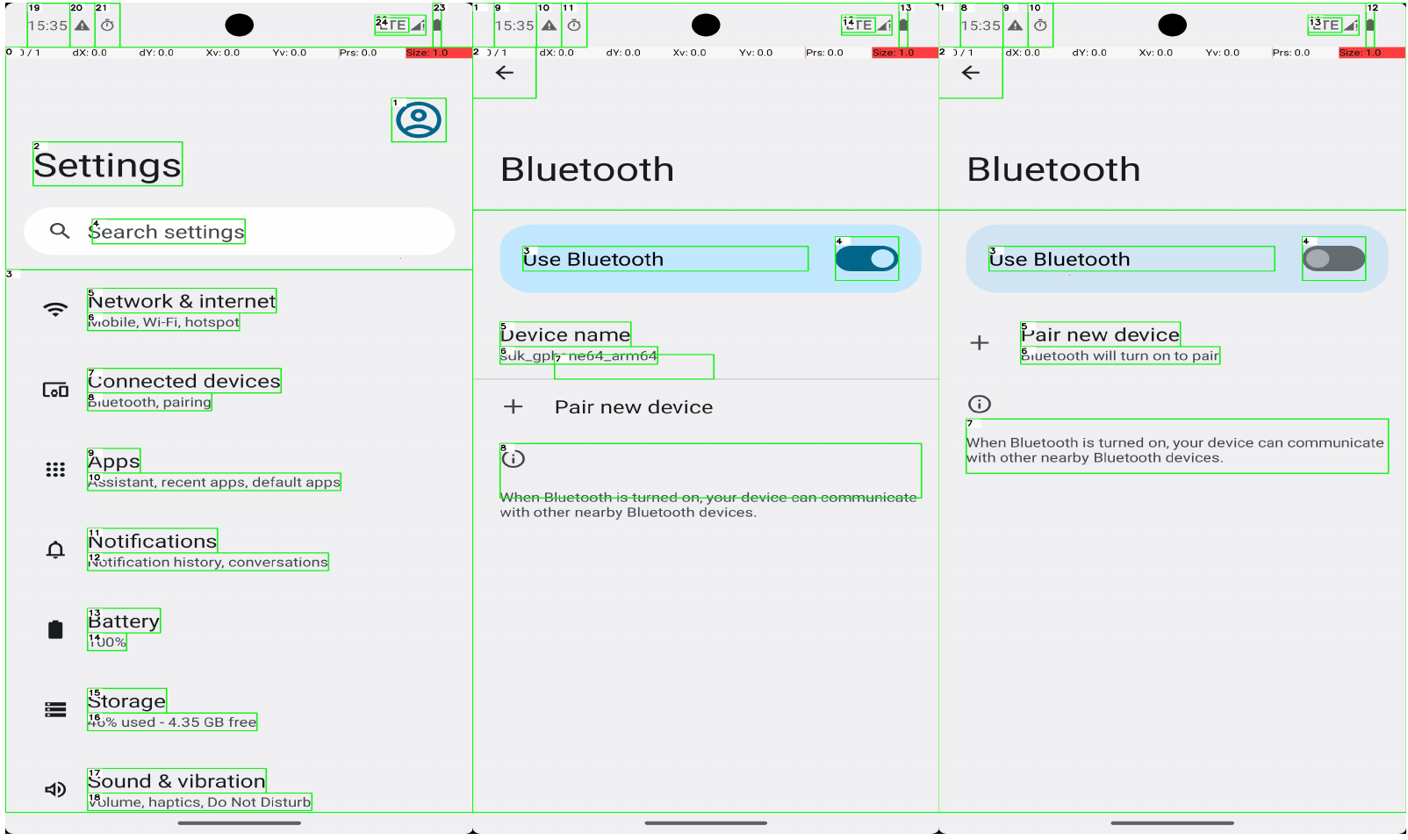}
    \vspace{-4mm}
   \caption{Turn Bluetooth Off}
   \vspace{-4mm}
   \label{fig:demo6}
\end{figure*}

The targeted task to be executed is "Turn Bluetooth off." The actual execution steps can be summarized as follows:
\begin{itemize}
    \item Step1: Open "Settings" App
    \item Step2: Click "Connected devices"
    \item Step3: Click "Connection preferences"
    \item Step4: Click "Bluetooth"
    \item Step5: Click "Bluetooth switch"
\end{itemize}
As Figure~\ref{fig:demo6} shows, we end up on the screenshot with the Bluetooth. We turn Bluetooth off, effectively completing the target task.

\subsection{Play Music Playlist}
\begin{figure*}[h]
     \centering
   \includegraphics[page=2,width=0.95\linewidth ]{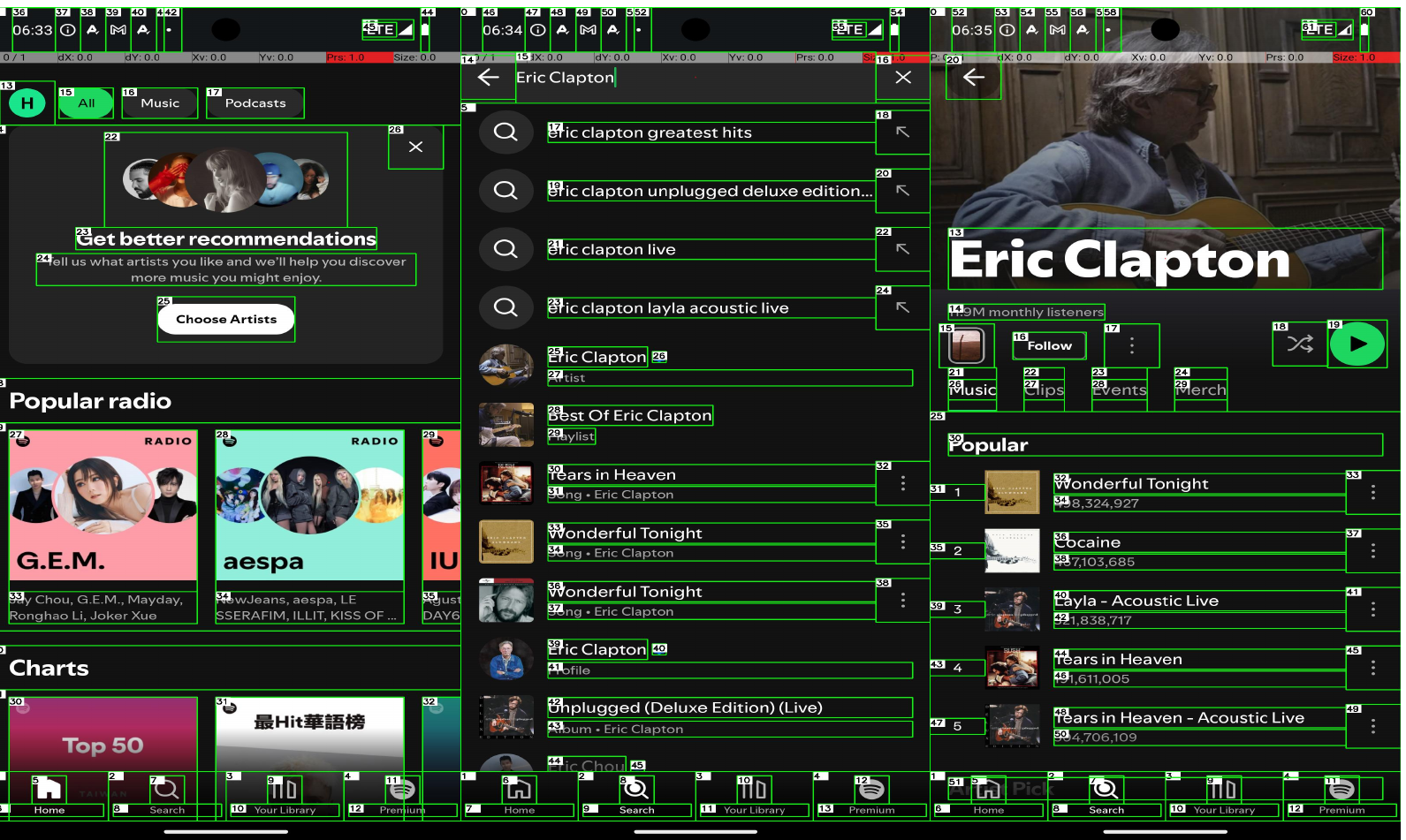}
    \vspace{-4mm}
   \caption{Play Music Playlist}
   \vspace{4mm}
   \label{fig:demo7}
\end{figure*}

The targeted task is "Search for the artist Eric Clapton, enter his artist page, and play his playlist." The actual execution steps can be summarized as follows:
\begin{itemize}
    \item Step1: Open "Spotify" App
    \item Step2: Click "Search"
    \item Step3: Click SearchBar
    \item Step4: Type "Eric Clapton"
    \item Step5: Click "Play"
\end{itemize}
As Figure \ref{fig:demo7} shows, we end up on the screenshot with the playlist. We play Eric Clapton's playlist, effectively completing the target task.

\subsection{Join a community}
\begin{figure*}[h]
     \centering
   \includegraphics[page=1,width=0.95\linewidth ]{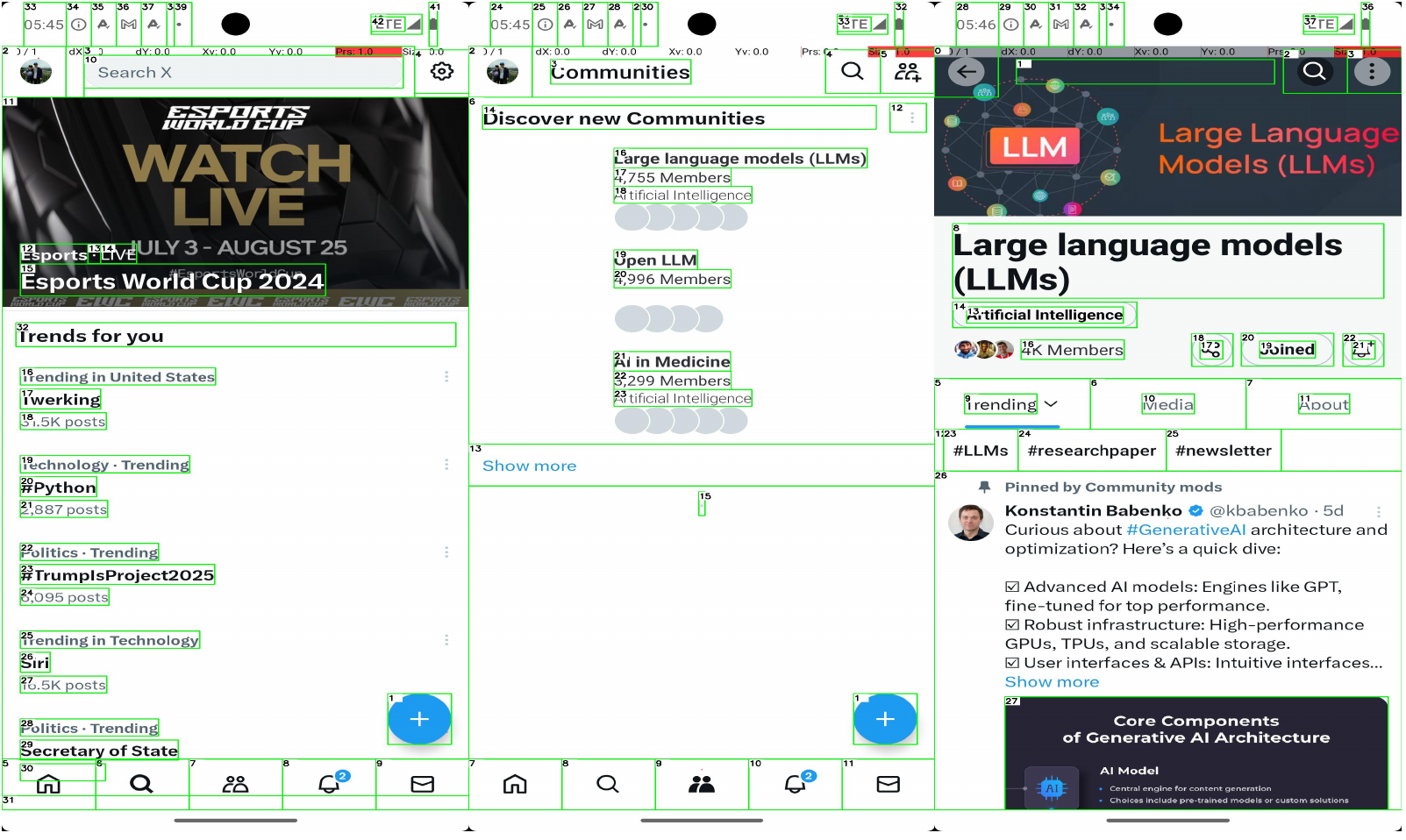}
    \vspace{-4mm}
   \caption{Join a community}
   \vspace{-4mm}
   \label{fig:demo8}
\end{figure*}

The targeted task is "Enter a community page and ask to join." The actual execution steps can be summarized as follows:
\begin{itemize}
    \item Step1: Open "Twitter" App
    \item Step2: Click "communities"
    \item Step3: Click a community
    \item Step4: Click "Join"
    \item Step5: Click "Agree and join"
\end{itemize}
As Figure~\ref{fig:demo8} shows, we end up on the screenshot with a community. We join the large language model community, effectively completing the target task.

\subsection{Math Counting Error}
\begin{figure*}[h]
     \centering
   \includegraphics[page=1,width=0.95\linewidth ]{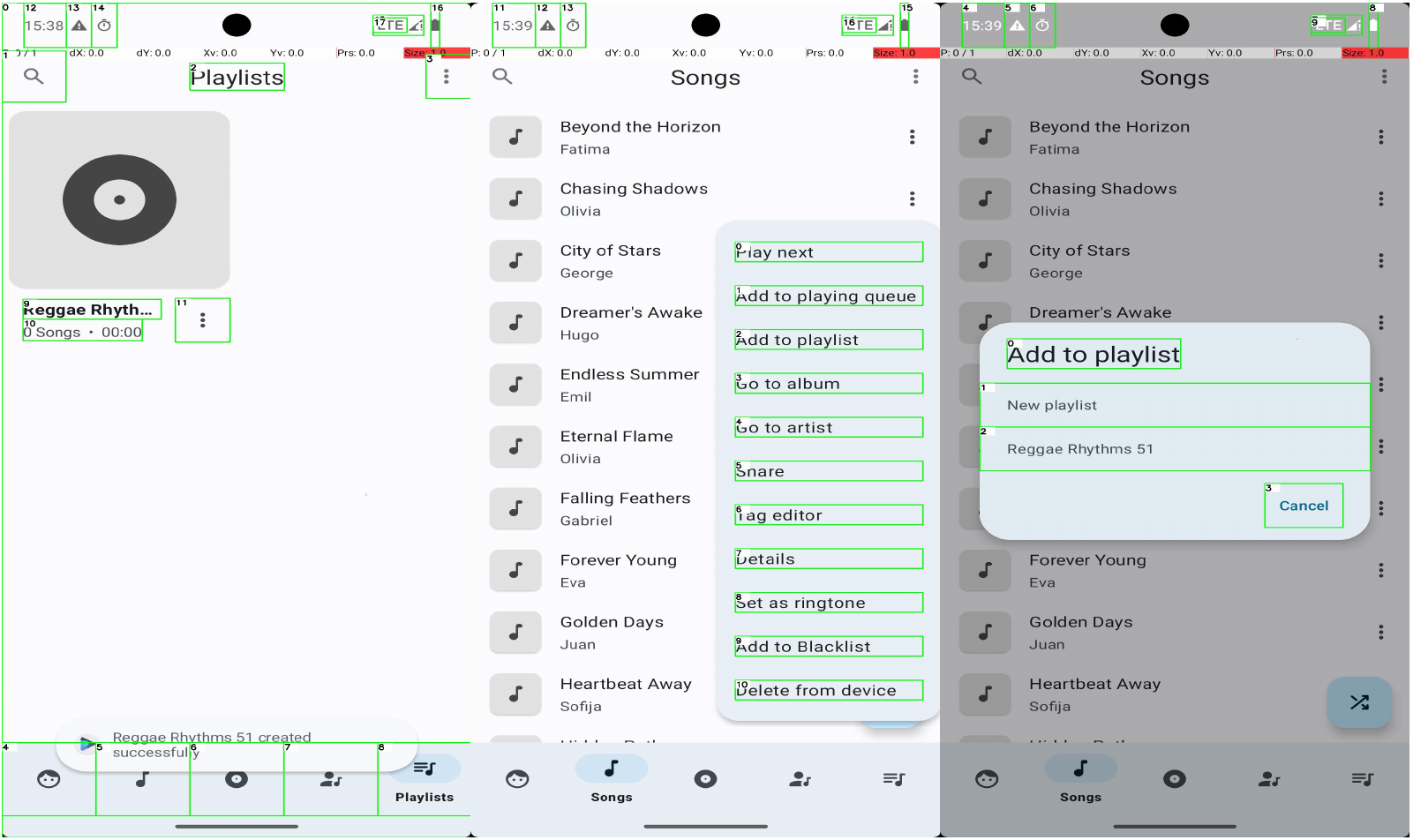}
    \vspace{-4mm}
   \caption{Math Counting Error}
   \vspace{4mm}
   \label{fig:demo9}
\end{figure*}

The targeted task to be executed is "Create a playlist in Retro Music titled "Reggae Rhythms 51" with a duration between 45 and 50 minutes using the provided songs". The actual execution steps can be summarized as follows:
\begin{itemize}
    \item Step1: Open app "Retro Music" App
    \item Step2: Click "Playlist"
    \item Step3: Click "More Options"
    \item Step4: Click "New Playlist"
    \item Step5: Type "Reggae Rhythms 51"
    \item Step6: Click "Create"
    \item Step7: Click "Songs"
    \item Step8: Long Press "Golden Days"
    \item Step9: Click "Add to playlist"
    \item Step10: Click "Reggae Rhythms 51"
    \item Step11: Long Press "Beyond the Horizon"
    \item Step12: Click "Add to playlist"
    \item Step13: Click "Reggae Rhythms 51"
    \item Step14: Long Press "Chasing Shadows"
    \item Step15: . . .
\end{itemize}
As Figure~\ref{fig:demo9} shows, since the model cannot calculate the song's duration, it will consider that the task has not been completed and continue adding songs, resulting in task failure.

\subsection{Memorization Error}
\begin{figure*}[h]
     \centering
   \includegraphics[page=1,width=0.95\linewidth ]{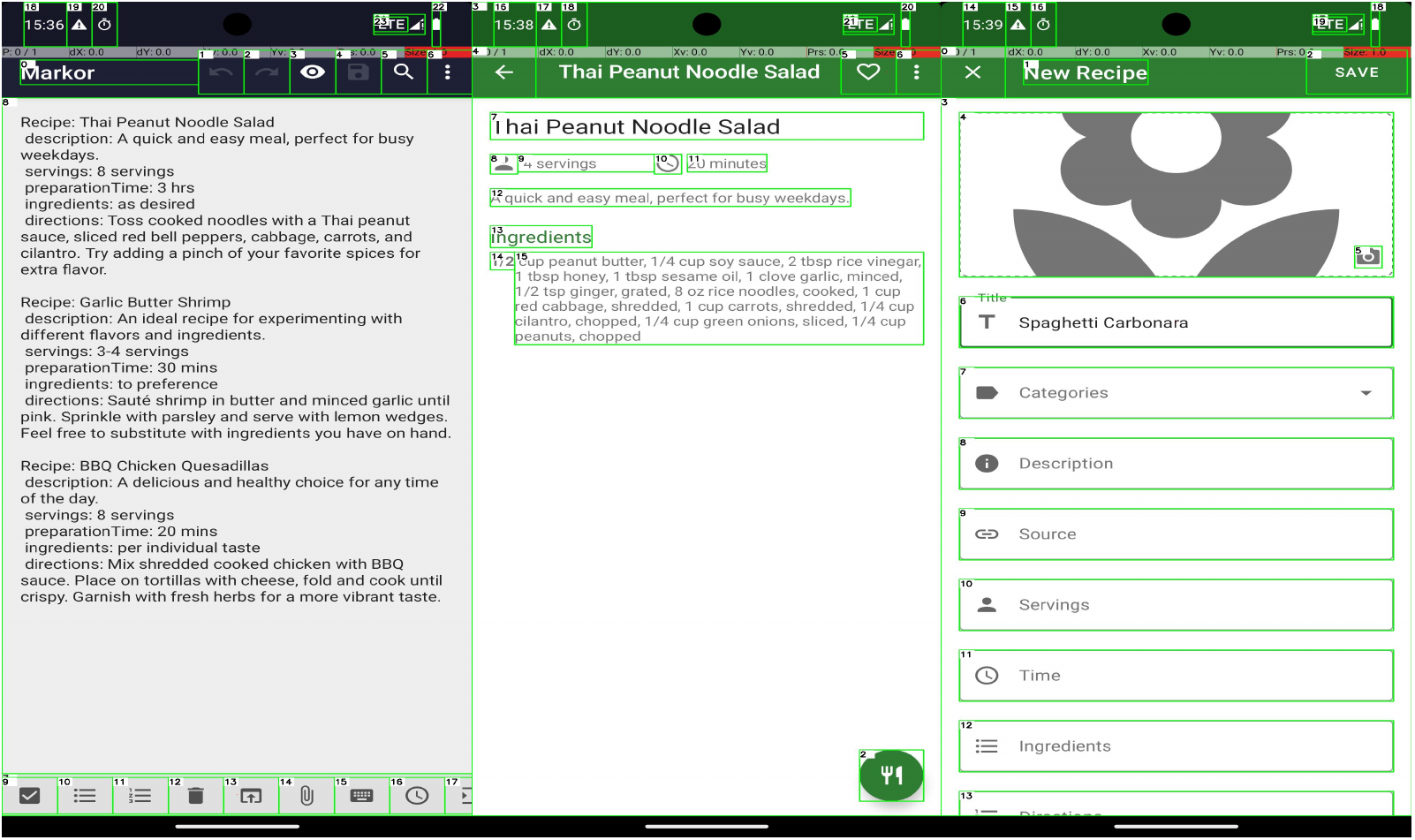}
    \vspace{-4mm}
   \caption{Memorization Error}
   \vspace{-4mm}
   \label{fig:demo10}
\end{figure*}

The targeted task is "Add the recipes from recipes.txt in Markor to the Broccoli recipe app." The actual execution steps can be summarized as follows:
\begin{itemize}
    \item Step1: Open "Markor" App
    \item Step2: Click "recipes.txt"
    \item Step3: Quote the content
    \item Step4: Navigate Home
    \item Step5: Open "Broccolirecipe" APP
    \item Step6: Click "Add"
    \item Step7: Type Title content
    \item Step8: Type Description content
    \item Step9: Type Serving content
    \item Step10: Type Time content
    \item Step11: Type Ingredients content
    \item Step12: Type "Save"
    \item Step13: Click "Add"
    \item Step14: Type Title content
    \item Step15: Fail
\end{itemize}
As Figure~\ref{fig:demo10} shows, we end up on the screenshot with Pages that have not been filled out. This is because the number of steps of the task exceeds the maximum number of steps the model can carry. 

\subsection{Vision Error}
\begin{figure*}[h]
     \centering
   \includegraphics[page=1,width=0.95\linewidth ]{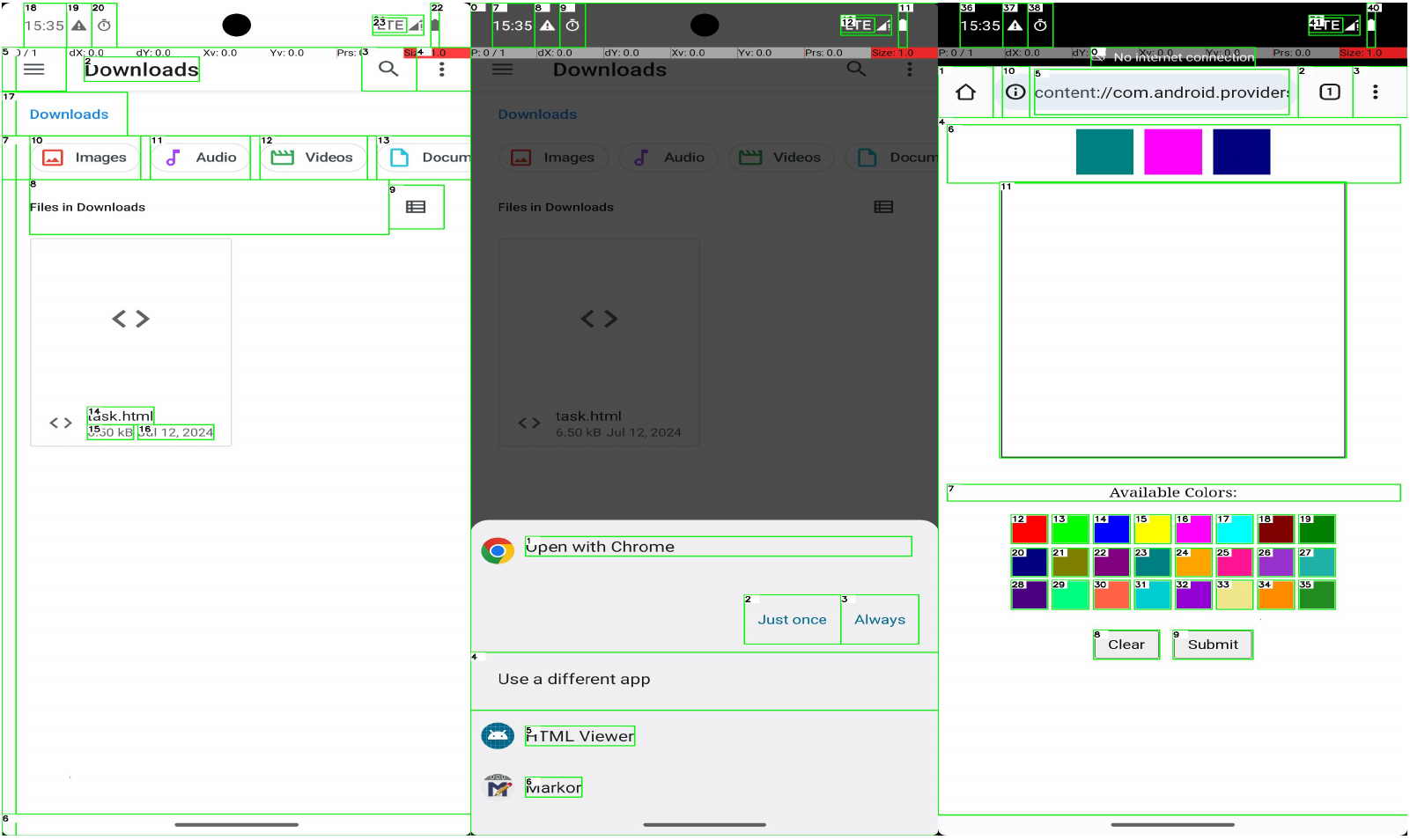}
    \vspace{-4mm}
   \caption{Vision Error}
   \vspace{-4mm}
   \label{fig:demo11}
\end{figure*}

The targeted task to be executed is "Open the file task.html in Downloads in the file manager; when prompted, open it with Chrome. Then create a drawing using the three colors shown at the top and hit submit". The actual execution steps can be summarized as follows:
\begin{itemize}
    \item Step1: Open "Files" App
    \item Step2: Click "task.html"
    \item Step3: Click "Open with Chrome"
    \item Step4: Click "Colors"
    \item Step5: Fail
\end{itemize}
As Figure~\ref{fig:demo11} shows, we end up with an empty drawing on the screenshot. This is because the model has no visual information, cannot obtain specific color information, and fails to perform the task of selecting colors. 

\end{document}